\pdfoutput=1

\documentclass[11pt]{article}

\usepackage[final]{acl}

\usepackage{times}
\usepackage{latexsym}

\usepackage[T1]{fontenc}

\usepackage[utf8]{inputenc}

\usepackage{microtype}

\usepackage{inconsolata}

\usepackage{amsmath,amsfonts,bm}

\def\eqref#1{equation~\ref{#1}}

\def\1{\bm{1}}

\DeclareMathAlphabet{\mathsfit}{\encodingdefault}{\sfdefault}{m}{sl}
\SetMathAlphabet{\mathsfit}{bold}{\encodingdefault}{\sfdefault}{bx}{n}

\usepackage{hyperref}
\usepackage{url}

\usepackage{color}
\usepackage{colortbl}
\usepackage{multirow}
\usepackage{tcolorbox}
\usepackage{booktabs}
\usepackage{subfig}
\usepackage{algorithm, algpseudocode}
\usepackage{wrapfig,lipsum}
\usepackage[russian,english]{babel}

\newcommand{\sysname}{{LongLLMLingua}}
\newcommand{\ie}{\textit{i.e.}}
\newcommand{\eg}{\textit{e.g.}}

\newtcolorbox[list inside=prompt]{prompt}[1][]{
    coltitle=white,
    boxsep=5pt,
    left=0pt,
    right=-1pt,
    top=0pt,
    bottom=0pt,
    boxrule=1pt,
    #1,
}

\title{\textit{\sysname{}}: Accelerating and Enhancing LLMs in Long Context Scenarios via Prompt Compression}

\author{Huiqiang Jiang,
Qianhui Wu,
Xufang Luo,\\
\bf Dongsheng Li,
Chin-Yew Lin,
Yuqing Yang,
Lili Qiu \\
Microsoft Corporation \\
\tt \{hjiang,qianhuiwu,xufluo,dongsli,cyl,yuqyang,liliqiu\}@microsoft.com \\
}

\begin{document}
\maketitle
\begin{abstract}

In long context scenarios, large language models (LLMs) face three main challenges: higher computational cost, performance reduction, and position bias.
Research indicates that LLM performance hinges on the density and position of key information in the input prompt.
Inspired by these findings, we propose \sysname{} for prompt compression towards improving LLMs' perception of the key information to simultaneously address the three challenges.
Our extensive evaluation across various long context scenarios demonstrates that \sysname{} not only enhances performance but also significantly reduces costs and latency.
For instance, in the NaturalQuestions benchmark, \sysname{} boosts performance by up to 21.4\% with around 4x fewer tokens in GPT-3.5-Turbo, leading to substantial cost savings.
It achieves a 94.0\% cost reduction in the LooGLE benchmark.
Moreover, when compressing prompts of about 10k tokens at ratios of 2x-6x, \sysname{} can accelerate end-to-end latency by 1.4x-2.6x.\footnote{Access our code at \url{https://aka.ms/LongLLMLingua}.}

\end{abstract}

\section{Introduction}
\label{sec:introduction}

Large language models (LLMs) have revolutionized user-oriented language technologies and are serving as crucial components in more and more applications.
Carefully designing prompts is necessary to achieve better performance in specific downstream tasks.
The commonly used technologies such as In-Context Learning (ICL)~\citep{min-etal-2022-metaicl,dong2022survey}, %
Retrieval Augment Generation (RAG)~\citep{lewis2020retrieval,asai2024selfrag}, and Multi-turn Agent~\citep{shen2024hugginggpt,park2023generative, wu2023autogen} are driving prompts to be increasingly longer, even reaching thousands of tokens.
Scenarios such as multi-document question answering, code completion, and document summarization also necessitate the processing of long contexts.

\begin{figure*}[hbt]
  \centering
  \subfloat[Performance v.s. Document Number]{
    \label{sfig:unrelated}
    \includegraphics[height=0.7\columnwidth]{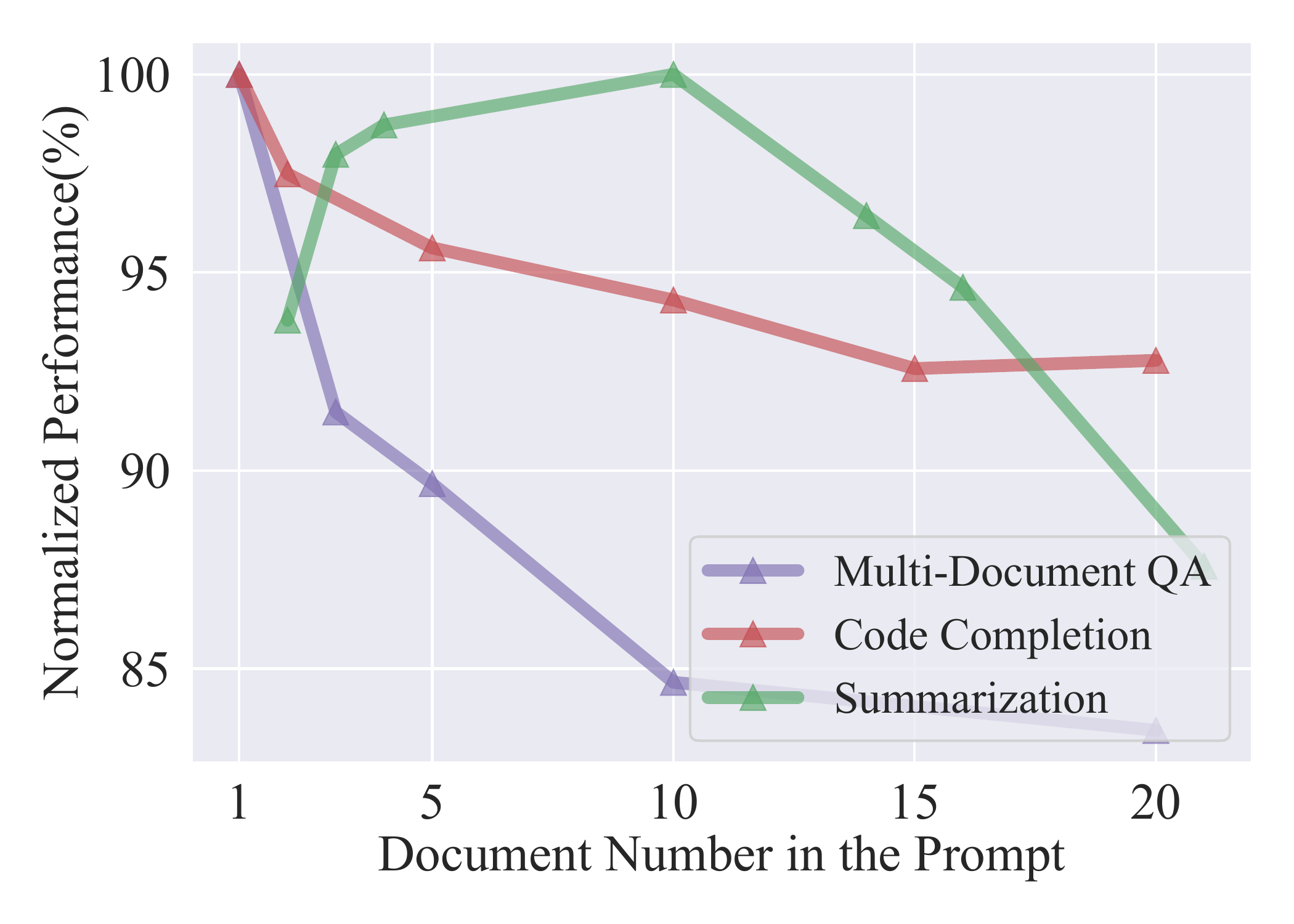}}
    \hspace{0.1em}
  \subfloat[Performance v.s. Key Information Position]{
    \label{sfig:compressed}
    \includegraphics[height=0.7\columnwidth]{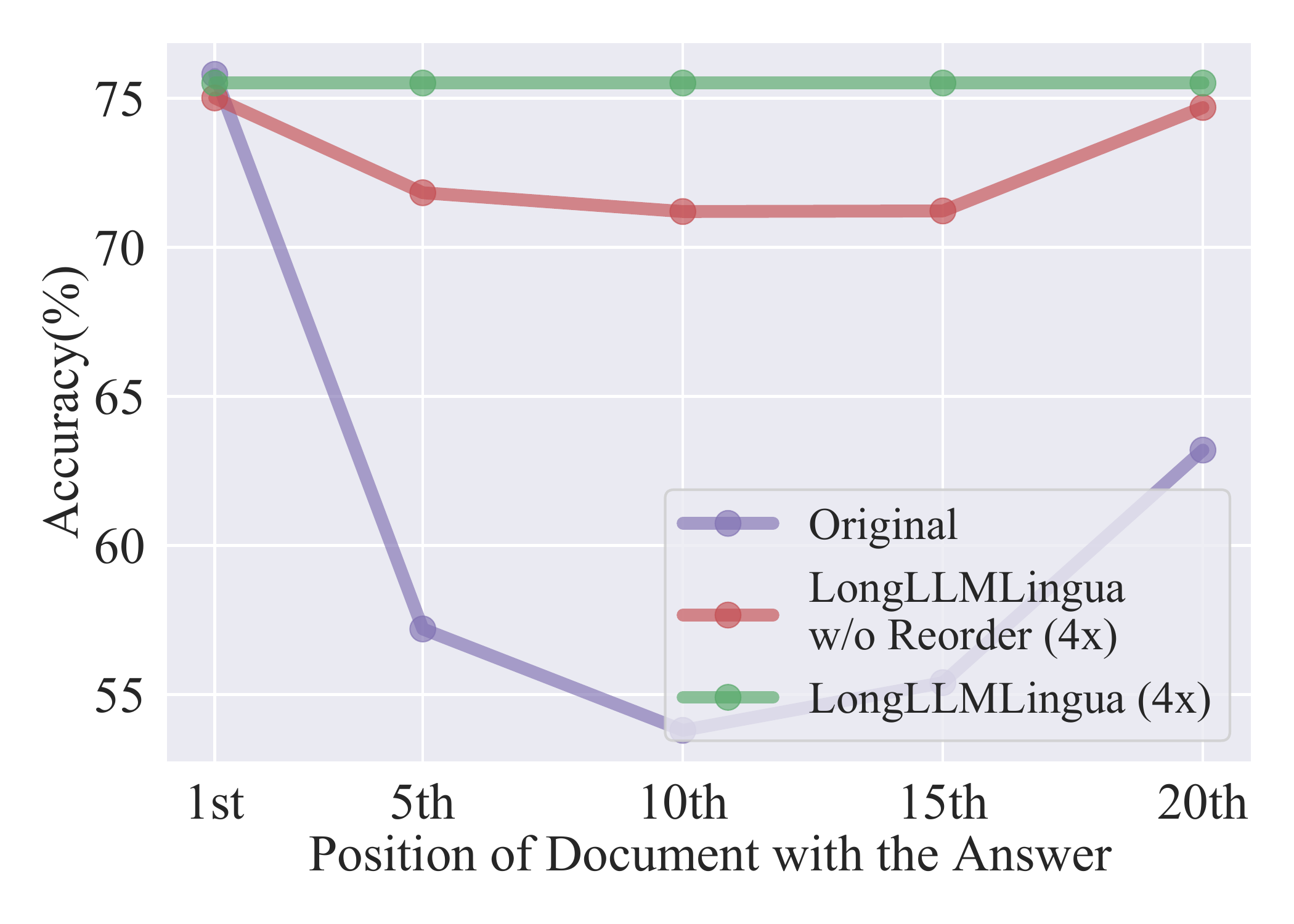}}
  \caption{(a) LLMs' performance in downstream tasks decreases with increased noise in prompts.
  In this case, we keep $k$ most relevant documents/paragraphs based on the ground-truth or LongLLMLingua $r_k$.
  A larger $k$ implies more noise introduced into the prompt.
  To improve the key information density in the prompt, we present question-aware coarse-to-fine compression.
  (b) LLMs' ability to capture the relevant information depends on their positions in the prompt. To reduce information loss in the middle, we introduce a document reordering mechanism.
  }
  \label{fig:motivations}
\end{figure*}

There are three main challenges when LLMs are used in long context scenarios:
(1) Higher computational costs, encompassing both financial and latency expenses.
(2) Longer prompts introduce irrelevant and redundant information, which can weaken LLMs' performance~\citep{shi2023large}, as illustrated in Figure~\ref{sfig:unrelated}.
(3) LLMs exhibit position bias~\cite{Greg_Needle_2023}, also known as the "lost in the middle" issue~\citep{liu2023lost}, suggesting that the placement of key information within the prompt significantly affects LLMs' performance. This is demonstrated by the purple curve in Figure \ref{sfig:compressed}.

Inspired by these observations, we propose \textit{\sysname{}} to address the three challenges.
Specifically, 
we use 
LLMLingua~\citep{anonymous2023llmlingua} as the backbone for prompt compression to address the first challenge, \ie, reduce cost and latency.
However, in the case of long contexts, the distribution of question-relevant key information
in the prompt is generally dynamic and sparse.
Existing prompt compression methods like LLMLingua~\citep{anonymous2023llmlingua} and Selective-Context~\citep{li2023unlocking} that often fail to consider question during compression, resulting in retention of excessive noise and decreased performance.
\sysname{} aims to improve LLMs' perception of key information pertinent to the question, thereby overcoming the noise and position bias issues in long contexts, shown in Figure~\ref{sfig:compressed}.
The underlying principle of \sysname{} is that small LM are inherently capable of capturing the distribution of key information relevant to a given question.

Our main contributions are five-fold:
(1) We propose a question-aware coarse-to-fine compression method to improve the key information density in the prompt (Sec. \ref{subsec:question_aware});
(2) We introduce a document reordering strategy to minimize position bias in LLMs. (Sec. \ref{subsec:doc_reorder});
(3) We establish dynamic compression ratios for precise control between coarse and fine compression levels (Sec. \ref{subsec:dynamic_compression_ratio});
(4) We propose a post-compression subsequence recovery strategy to improve the integrity of the key information (\ref{subsec:recover}).
(5) We evaluate \sysname{} across five benchmarks, \ie, NaturalQuestions~\citep{liu2023lost}, LongBench~\citep{bai2023longbench}, ZeroSCROLLS~\citep{shaham2023zeroscrolls}, MuSicQue~\citep{trivedi2021musique}, and LooGLE~\citep{li2023loogle}, covering a variety of long context scenarios.  %
Experimental results reveal that \sysname{}'s compressed prompts outperform original prompts in terms of performance, cost efficiency, and system latency.

\section{Problem Formulation}

Following LLMLingua~\citep{anonymous2023llmlingua}, we use $\mathbf{x}=(\mathbf{x}^{\text{ins}}, \mathbf{x}^{\text{doc}}_1, \cdots, \mathbf{x}^{\text{doc}}_K, \mathbf{x}^{\text{que}})$ to represent a prompt, including the instruction $\mathbf{x}^{\text{ins}}$, $K$ documents $\mathbf{x}^{\text{doc}}_i$, and the question $\mathbf{x}^{\text{que}}$.
However, this definition can be adjusted for specific scenarios.
The objective of a prompt compression system can be formulated as:
\begin{equation}
\min_{\widetilde{\mathbf{x}}} D_{\phi}\left(\mathbf{y}, \widetilde{\mathbf{y}} \right) + \lambda\lVert \widetilde{\mathbf{x}}\rVert_0,
\end{equation}
where $\widetilde{\mathbf{x}}$ represents the compressed prompt, a token-level subsequence of $\mathbf{x}$. 
$\mathbf{y}$ and $\widetilde{\mathbf{y}}$ represent the LLM-generated results from $\mathbf{x}$ and $\widetilde{\mathbf{x}}$, respectively. 
$D_{\phi}$ measures the distance function, such as KL divergence.
$\lambda$ serves as a hyper-parameter balancing the compression ratio. Additionally, this study explores a permutation operation space over the $K$ documents $(\mathbf{x}^{\text{doc}}_1, \cdots, \mathbf{x}^{\text{doc}}_K)$ for joint optimization.

\begin{figure*}[tb]
    \centering
    \includegraphics[width=\linewidth]{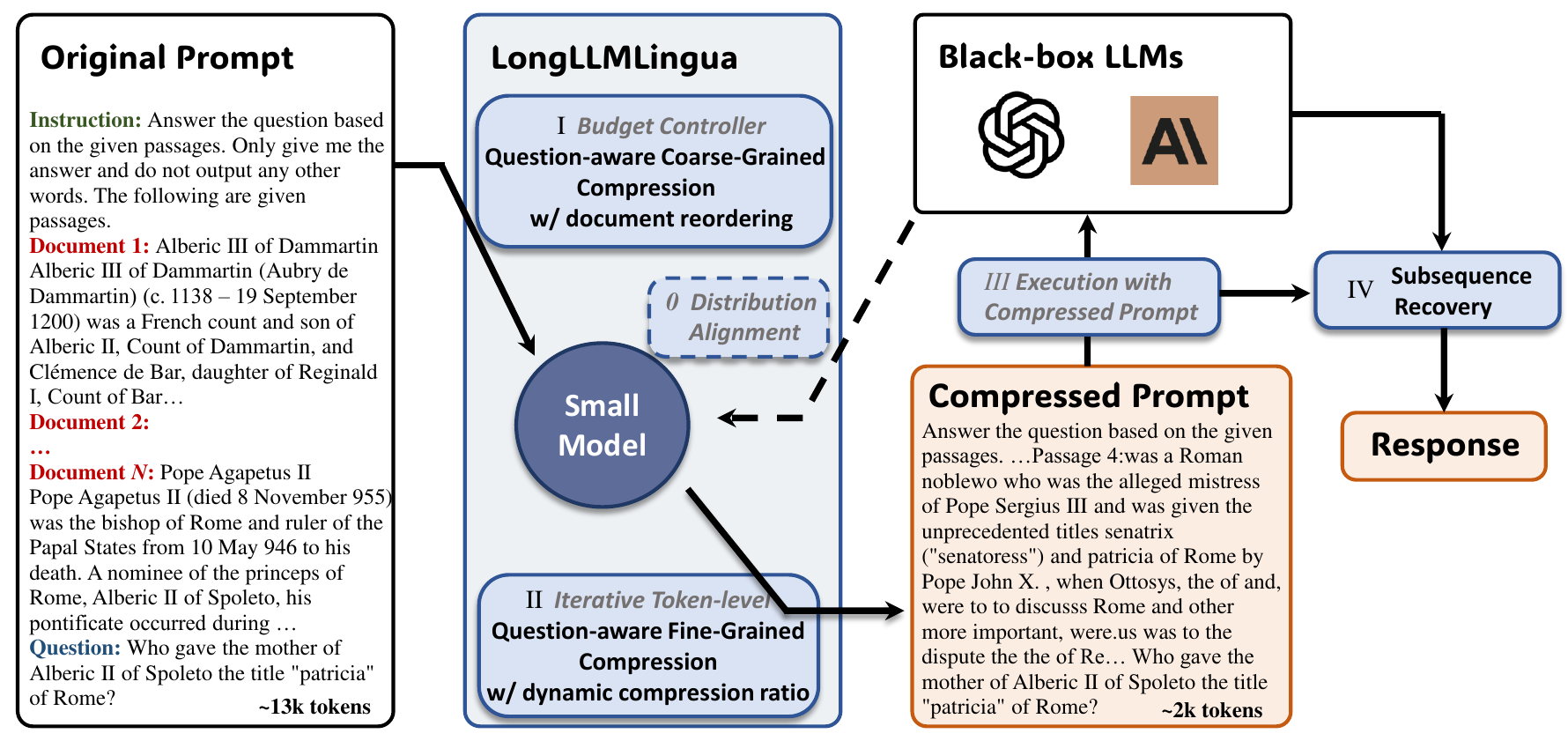}
    \caption{Framework of \textit{\sysname{}}. \textcolor{gray}{Gray \textit{Italic}} content: As in LLMLingua.}
    \label{fig:framework}
\end{figure*}

\section{Preliminary: LLMLingua}
LLMLingua~\citep{anonymous2023llmlingua} utilizes a small language model $\mathcal{M}_{S}$ to evaluate the perplexity of each prompt token, removing those with lower perplexities. This method is premised on the idea that tokens with lower perplexities have a negligible effect on the language model's overall entropy gain, implying their removal slightly impacts the LLMs' contextual understanding. This process is viewed as an application of "LM is Compression"~\citep{deletang2023language}.
LLMLingua include three key components: budget controller, iterative token-level prompt compression, and distribution alignment, highlighted by italic text in Figure~\ref{fig:framework}. The budget controller assigns varying compression ratios to different parts of the prompt (i.e., instruction, demonstrations, question), implementing coarse-level prompt compression. Subsequent steps involve dividing intermediate results into segments and applying token-level compression iteratively, where each token's perplexity based on preceding compressed segments.
To aware different target LLMs, LLMLingua fine-tunes $\mathcal{M}_{S}$ using data from the target LLM.

\section{LongLLMLingua} %

LongLLMLingua builds on LLMLingua to better compress prompts in long context scenorias. It tackles three main issues in handling lengthy contexts, as introduced in Sec.~\ref{sec:introduction}. This approach focuses on making LLMs more effective at recognizing key information related to the question in the prompt. It encompasses three perspectives and further incorporates a subsequence recovery strategy, as shown in Figure~\ref{fig:framework}, to enhance the accuracy and reliability of the information provided to users.
In this section, we detail how each part of LongLLMLingua works to improve the LLMs deal with long context.

\subsection{How to improve key information density in the prompt?}
\label{subsec:question_aware}

\paragraph{Question-Aware Coarse-Grained Compression}

\begin{figure*}[htb]
  \centering
  \subfloat[Recall Distribution]{
    \label{sfig:recall}
    \includegraphics[height=0.7\columnwidth]{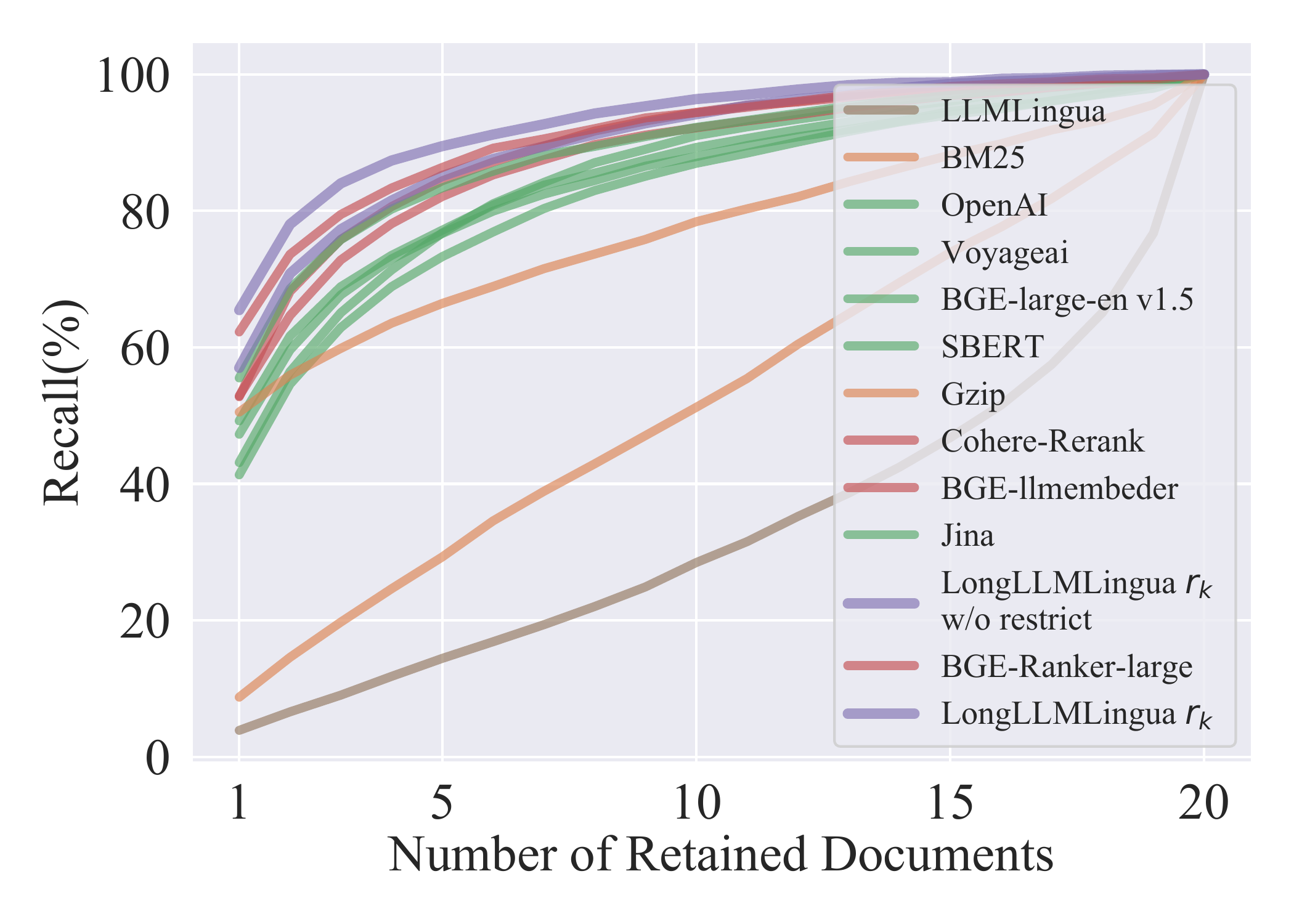}}
    \hspace{1em}
  \subfloat[Perplexity Distribution (5th)]{
    \label{sfig:ppl}
    \includegraphics[height=0.7\columnwidth]{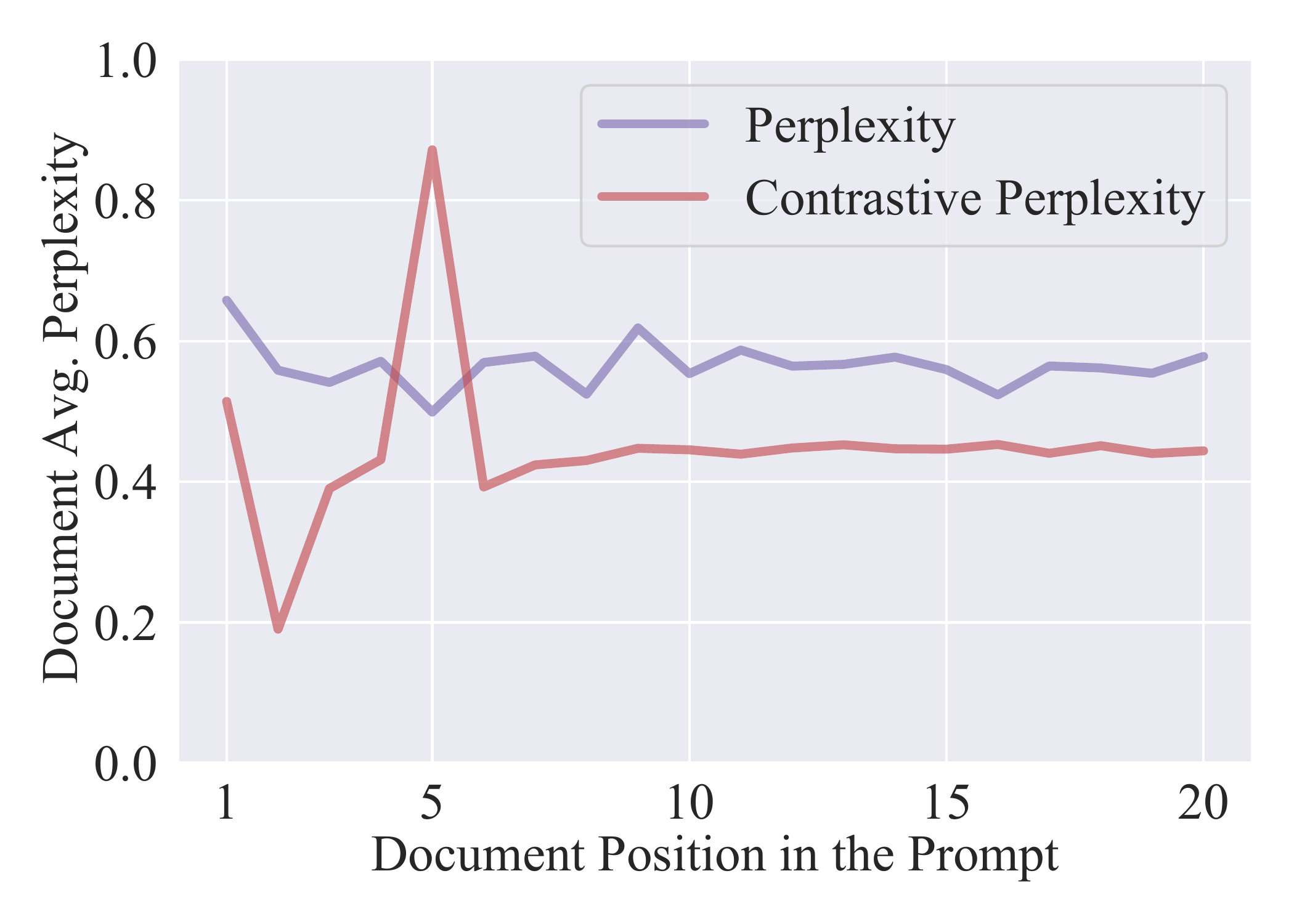}}
  \caption{(a) Comparison of recall on NaturalQuestions Multi-documemnt QA dataset, which increases from top to bottom in terms of Recall@1. Different colors represent different types of methods. Among them, yellow represents traditional relevance methods, green signifies embedding-based methods, and red denotes rerank-based methods. (b) Comparison between perplexities and contrastive perplexities of tokens in the prompt from Multi-documemnt QA dataset.
  The document containing the ground-truth information is located in the 5th position.
  More results on position can be found in the Appendix~\ref{subsec:empirical}.
  }
  \label{fig:method_motivations}
\end{figure*}

In coarse-grained compression, we aim to figure out a metric $r_k$ to evaluate the importance of each document $\mathbf{x}^{\text{doc}}_k=\{x_{k,i}^{\text{doc}}\}_{i=1}^{N_k}$, where $N_k$ is the number of tokens in $\mathbf{x}^{\text{doc}}_k$.
We only keep $\mathbf{x}^{\text{doc}}_k$ with higher $r_k$ as the intermediate compressed results.
One approach to improve key information density in the compressed prompts is to calculate document-level perplexity conditioned on the question $p(\mathbf{x}_{k}^{\text{doc}}|\mathbf{x}^{\text{que}})$.
However, this method may not be effective because documents often contain a significant amount of irrelevant information. Even when conditioned on $\mathbf{x}^{\text{que}}$, the perplexity scores computed for entire documents may not be sufficiently distinct, rendering them an inadequate metric for document-level compression.

We propose to use the perplexity of the question $\mathbf{x}^{\text{que}}$ conditioned on different contexts $\mathbf{x}^{\text{doc}}_k$ $p(\mathbf{x}^{\text{que}}|\mathbf{x}_{k}^{\text{doc}})$ to represent the association between them.
We also append a restrictive statement\footnote{Specifically, "\textit{We can get the answer to this question in the given documents}".
} $\mathbf{x}^{\text{restrict}}$
after $\mathbf{x}^{\text{que}}$ to strengthen the interconnection of $\mathbf{x}^{\text{que}}$ and $\mathbf{x}^{\text{doc}}_k$.
It can be regarded as a regularization term that mitigates the impact of hallucinations.
This can be formulated as:
\begin{equation}
\begin{aligned}
    r_k = -\frac{1}{N_c}\sum_{i}^{N_c} \log p(x^{\text{que}, \text{restrict}}_i|\mathbf{x}_{k}^{\text{doc}})&, \\k \in \{1,2,\cdots,K\}&,
\end{aligned}
\label{eq:longllmlingua_coarse}
\end{equation}
where $x^{\text{que}, \text{restrict}}_i$ is the $i$-th token in the concatenated sequence of $\mathbf{x}^{\text{que}}$ and $\mathbf{x}^{\text{restrict}}$ and $N_c$ in the number of tokens.

Figure~\ref{sfig:recall} displays the recall distribution of different retrieval methods, including traditional relevance methos (BM25, Gzip~\citep{jiang-etal-2023-low}), embedding-based methods (OpenAI-embedding, Voyageai\footnote{https://www.voyageai.com/}, BGE-large-en v1.5~\citep{bge_embedding}, Sentence-BERT~\citep{reimers-2019-sentence-bert}, Jina~\citep{günther2023jina}), and reranker methods (Cohere-Rerank\footnote{https://cohere.com/rerank}, BGE-llmembeder, BGE-Ranker-large), which demonstrates that our coarse-level compression approach achieves the highest recall with different numbers of retained documents, suggesting that it preserves the most key information from the contexts in the compressed results.

\paragraph{Question-Aware Fine-Grained Compression}
In fine-grained compression, we assess the importance of each token in the instruction $\mathbf{x}^{\text{ins}}$, the question $\mathbf{x}^{\text{que}}$, and $K'$ documents $\{\mathbf{x}^{\text{doc}}_i\}_{i=1}^{K'}$ retained after coarse-grained compression.
We incorporate the iterative compression mechanism following LLMLingua and directly calculate token perplexities to compress $\mathbf{x}^{\text{ins}}$ and $\mathbf{x}^{\text{que}}$. 
In this section, we investigate how to make the fine-grained token-level compression over $\{\mathbf{x}^{\text{doc}}_k\}_{k=1}^{K'}$ aware of the question $\mathbf{x}^{\text{que}}$, so that the compressed results could contain more question-relevant key information.

A straightforward solution for the awareness of $\mathbf{x}^{\text{que}}$ is to simply concatenate it at the beginning of the whole context.
However, this will result in low perplexities of relevant tokens in the context following the condition of question $\mathbf{x}^{\text{que}}$, further reducing their differentiation from other tokens.

In this paper, we propose \textit{contrastive perplexity}, \ie, the distribution shift caused by the condition of the question, to represent the association between the token and the question.
The contrastive perplexity based importance metric $s_i$ for each token $x_i$ in $\{\mathbf{x}^{\text{doc}}_k\}_{k=1}^{K'}$ can be formulated as:
\begin{equation}
    s_i = \text{perplexity}(x_i|x_{<i}) - \text{perplexity}(x_i|x^{\text{que}}, x_{<i}).
    \label{eq:token_condition}
\end{equation}

Additionally, we provide the derivation of its mathematical significance in the Appendix~\ref{sec:derivation}, concluding that it is equivalent to conditional pointwise mutual information~\citep{church1990word}.

Figure~\ref{sfig:ppl} illustrates the difference between perplexities and contrastive perplexities.
The distribution of perplexities appears random, making it challenging to extract information related to the question. However, tokens with high contrastive perplexities tend to cluster near the ground-truth document, which contains information relevant to the question.
This suggests that the proposed contrastive perplexity can better distinguish tokens relevant to the question, thus improving the key information density in the compressed results.

\subsection{How to reduce information loss in the middle?}
\label{subsec:doc_reorder}

\begin{figure*}[htb]
    \centering
    \resizebox{1.99\columnwidth}{!}{
    \includegraphics[width=1\linewidth]{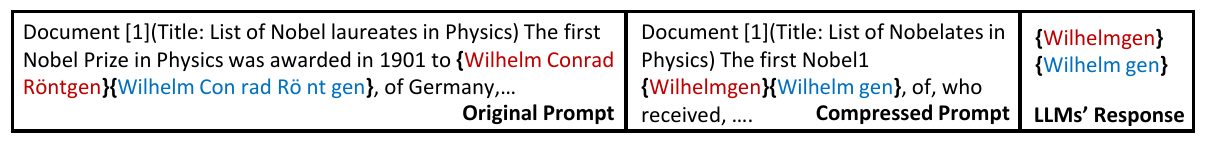}
    }
    \caption{The example of Subsequence Recovery, the red text represents the original text, and the blue text is the result after using the LLaMA 2-7B tokenizer.}
    \label{fig:recovery_case}
\end{figure*}

As demonstrated in Figure \ref{sfig:compressed},
LLM achieves the highest performance when relevant information occurs at the beginning and significantly degrades if relevant information is located in the middle of long contexts.
After the coarse-grained compression, we have obtained a set of documents $\{\mathbf{x}^{\text{doc}}_k\}_{k=1}^{K'}$ with their corresponding importance scores $\{r_k\}_{k=1}^{K'}$ indicating their association with the question $\mathbf{x}^{\text{que}}$.
Therefore, we reorder documents using their importance scores to better leverage LLMs' information perception difference in positions:

\begin{equation}
\begin{aligned}
    (\mathbf{x}^{\text{ins}}, \mathbf{x}^{\text{doc}}_1, \cdots, \mathbf{x}^{\text{doc}}_{K'}, &\mathbf{x}^{\text{que}}) \stackrel{r_k}{\longrightarrow}\\ (\mathbf{x}^{\text{ins}}, &\mathbf{x}^{\text{doc}}_{r1}, \cdots, \mathbf{x}^{\text{doc}}_{rK'}, \mathbf{x}^{\text{que}})
    \label{eq:reorder}
\end{aligned}
\end{equation}

\subsection{How to achieve adaptive granular control during compression?}
\label{subsec:dynamic_compression_ratio}

In fine-grained compression, LLMLingua applies the same compression ratio over all documents obtained from budget controller. However, the key information density of different documents is different.
The more relevant to the question a document is, the more budget (\ie, lower compression ratio) we should allocate to it.
Therefore, we bridge coarse-grained compression to fine-grained compression and use the importance scores $\{r_k\}_{k=1}^{K'}$ obtained from coarse-grained compression to guide the budget allocation in fine-grained compression.
In this way, we can achieve adaptive granular control on the whole.

Specifically, we first determine the initial budget for the retained documents\footnote{In LLMLingua, it is $\tau^{\text{dems}}$ for demonstrations.} $\tau^{\text{doc}}$ using the budget controller of LLMLingua. 
During fine-grained compression, we follow the iterative token-level compression algorithm in LLMLingua but dynamically assign the compression budget $\tau_k^{\text{doc}}$ to each document $\mathbf{x}_k^{\text{doc}}$ according to the ranking index $I(r_k)$ (e.g., 0, 1) of its importance score from the coarse-grained compression.
In this paper, we employ a linear scheduler for the adaptive allocation. Budget of each token $x_i$ can be formulated as:
\begin{equation}
\begin{aligned}
    \tau_i &= \tau_k^{\text{doc}}, \quad \forall x_i \in \mathbf{x}^{\text{doc}}_k, \\
    \tau_k^{\text{doc}} &= \max(\min((1-\frac{2I(r_k)}{K'}) \delta\tau + \tau^{\text{doc}}, 1), 0), \\
\end{aligned}
\end{equation}
where $i$ and $k$ is the index of token and document, $K'$ denotes the number of documents, and $\delta\tau$ is a hyper-parameter that controls the overall budget for dynamic allocation.

\subsection{How to improve the integrity of key information?}
\label{subsec:recover}

During the generation process, LLMs tend to replicate entities found in the prompt, such as names, places, and organizations. Compressing these entities at the token level doesn't affect the LLMs' understanding of semantic content but can lead to errors in the generated content. 

Therefore, we propose a subsequence recovery method to restore the original content in LLMs' responses.
This method relies on the subsequence relationship among %
tokens in the original prompt, compressed prompt, and LLMs' response, as shown in Figure~\ref{fig:recovery_case}.

The overall procedure includes:  
i) Iterate through tokens $y_l$ in LLMs' response and select the longest substring $\bm{\widetilde{y}}_{\text{key},l}=\{y_l, y_{l+1}, ..., y_{r}\}$ that appears in the compressed prompt $\bm{\widetilde{x}}$.
ii) Find the maximum common shortest subsequence $\bm{{x}}_{i,j}=\{x_i, x_{i+1}, ..., x_{j}\}$ in the original prompt $\bm{x}$, corresponding to the representation $\bm{\widetilde{y}}_{\text{key},l}$ in the original prompt (accelerated using prefix trees or sequence automata).  
iii) Replace the matched tokens $\bm{\widetilde{y}}_{\text{key},l}$ in LLMs' response with the corresponding subsequence $\bm{{x}}_{i,j}$ from the original prompt.
For more details, please refer to Algorithm~\ref{alg:subsquence_recovery}.

\begin{algorithm}[htb]
    \small
	\caption{Token-level Subsquence Recovery Algorithm} 
    \textbf{Input}: The original prompt $\bm{x}$; the compressed prompt $\bm{\widetilde{x}}$; the generation response of LLMs $\bm{y}$.
	\begin{algorithmic}[1]
          \State Set the final response list $\bm{y}_\text{rec}=\phi$, the left token index of subsquence $l$ to 0.
         \While{$l < \bm{y}.len()$}
         \If { Substring ${y_l} \in \bm{\widetilde{x}}$}
         \State Find the longer substring $\bm{\widetilde{y}}_{\text{key},l}=\{y_l, y_{l+1},$ $ ..., y_{r}\} \in \bm{\widetilde{x}}$.
         \State Find the maximum common shortest subsequence $\bm{{x}}_{i,j}=\{x_i, x_{i+1}, ..., x_{j}\}$ in the original prompt $\bm{x}$.
         \State Add the subsequence $\bm{{x}}_{i,j}=\{x_i, x_{i+1}, ..., x_{j}\}$ to the response $\bm{y}_\text{rec}$.
         \State Set the left index $l$ to $r + 1$.
         \Else
         \State Add the token $y_l$ to the response $\bm{y}_\text{rec}$.
         \State Set the left index $l$ to $l + 1$.
         \EndIf
         \EndWhile
	\end{algorithmic} 
    \textbf{Output}: The final response list $\bm{y}_\text{rec}$.
    \label{alg:subsquence_recovery}
\end{algorithm}

\section{Experiments}

Here, we %
investigate: %
(1) How \textit{effective} is \sysname{}? %
(2) How \textit{efficient} is \sysname{}? %

\paragraph{Implementation details} In this paper, we use GPT-3.5-Turbo-0613\footnote{For experiments with original prompts exceeding 4k tokens, we utilize GPT-3.5-Turbo-16k-0613.} and LongChat-13B-16k as the target LLMs, both accessible via OpenAI\footnote{https://platform.openai.com} and HuggingFace\footnote{https://huggingface.co/lmsys/longchat-13b-16k}. To ensure stable and reproducible results, we employ greedy decoding and set the temperature to 0 in all experiments.
For the small language models used for compression, we apply LLaMA-2-7B-Chat\footnote{https://ai.meta.com/llama/}, which has been aligned by supervised fine-tuning and RLHF. We implement our approach with PyTorch 1.13.1 and HuggingFace Transformers.
We set up hyperparameters following LLMLingua except for the segment size used in iterative token-level compression set to 200 here.
More details are provided in Appendix~\ref{sec:experiment_details}.

\begin{table*}[tb]
    \small
    \centering
    \setlength{\tabcolsep}{1mm}
    \vspace{-2ex}
    \resizebox{2.05\columnwidth}{!}{
    \begin{tabular}{l|cccccc|cccccc|cc|cc}
    \toprule
        \multirow{2}{*}{Methods} &  \multicolumn{6}{@{}c}{{\bf GPT3.5-Turbo}} &  \multicolumn{6}{@{}c}{{\bf LongChat-13b}} & \multicolumn{2}{@{}c}{{\bf Length}} & \multicolumn{2}{@{}c}{{\bf Latency}}\\
        & 1st & 5th & 10th & 15th & 20th & Reorder & 1st & 5th & 10th & 15th & 20th & Reorder & Tokens & $1/\tau$ & Latency & Speedup \\
    \midrule
    \midrule
    \multicolumn{10}{@{}r}{{ \textit{2x constraint}}} \\
    \midrule
      \multicolumn{14}{@{}l}{{ \textit{Retrieval-based Methods}}} \\ 
    BM25 & 53.7 & 49.3 & 47.9 & 49.9 & 46.9 & 50.3 & 50.9 & 44.9 & 44.1 & 42.9 & 43.2 & 46.0 & 1,545 & 1.9x & 2.1 & 1.9x\\
    Gzip & 64.6 & 63.8 & 60.5 & 58.3 & 57.3 & 64.4 & 61.9 & 55.7 & 52.7 & 50.8 & 50.9 & 59.3 & 1,567 & 1.9x & 2.1 & 1.9x \\
    SBERT & 72.5 & 67.9 & 63.3 & 65.0 & 66.2 & 68.7 & 65.8 & 57.5 & 54.9 & 53.4 & 55.7 & 61.4 & 1,549 & 1.9x & 2.2 & 1.9x \\
    OpenAI & 73.0 & 65.6 & 66.5 & 65.4 & 65.5 & 69.9 & 65.9 & 57.5 & 56.2 & 54.2 & 55.7 & 61.7 & 1,550 & 1.9x & 4.9 & 0.8x \\
    LongLLMLingua $r_k$ & 73.9 & 67.7 & 68.7 & 66.0 & 65.6 & 74.3 & 68.5 & 59.1 & 56.8 & 55.3 & 56.9 & 65.2 & 1,548 & 1.9x & 2.3 & 1.8x \\
    \cmidrule (r){1-1}\cmidrule (lr){2-7} \cmidrule (lr){8-13} \cmidrule (lr){14-15} \cmidrule (lr){16-17}
    \multicolumn{7}{@{}l}{{ \textit{Compression-based Methods}}} \\
    Selective-Context & 45.4 & 39.0 & 33.8 & 33.5 & 41.5 & - & 53.2 & 26.3 & 25.4 & 24.2 & 33.3 & - & 1,478 & 2.0x & 7.4 & 0.6x \\
    LLMLingua & 39.7 & 39.5 & 40.4 & 37.1 & 42.3 & 41.5 & 38.7 & 37.3 & 35.7 & 34.1 & 37.5 & 37.1 & 1,410 & 2.1x & 2.8 & 1.5x \\
    \cmidrule (r){1-1}\cmidrule (lr){2-7} \cmidrule (lr){8-13} \cmidrule (lr){14-15} \cmidrule (lr){16-17}
    {\cellcolor[rgb]{0.925,0.957,1}}\textbf{LongLLMLingua} & {\cellcolor[rgb]{0.925,0.957,1}}\textbf{77.2} & {\cellcolor[rgb]{0.925,0.957,1}}\textbf{72.9} & {\cellcolor[rgb]{0.925,0.957,1}}\textbf{70.8} & {\cellcolor[rgb]{0.925,0.957,1}}\textbf{70.5} & {\cellcolor[rgb]{0.925,0.957,1}}\textbf{70.6} & {\cellcolor[rgb]{0.925,0.957,1}}\textbf{76.2} & {\cellcolor[rgb]{0.925,0.957,1}}\textbf{68.7} & {\cellcolor[rgb]{0.925,0.957,1}}\textbf{59.4} & {\cellcolor[rgb]{0.925,0.957,1}}\textbf{57.3} & {\cellcolor[rgb]{0.925,0.957,1}}\textbf{55.9} & {\cellcolor[rgb]{0.925,0.957,1}}\textbf{58.4} & {\cellcolor[rgb]{0.925,0.957,1}}\textbf{66.1} & {\cellcolor[rgb]{0.925,0.957,1}}1,429 & {\cellcolor[rgb]{0.925,0.957,1}}2.1x & {\cellcolor[rgb]{0.925,0.957,1}}2.9 & {\cellcolor[rgb]{0.925,0.957,1}}1.4x \\

    \midrule
    \midrule
    \multicolumn{10}{@{}r}{{ \textit{4x constraint}}} \\
    \midrule
      \multicolumn{15}{@{}l}{{ \textit{Retrieval-based Methods}}} \\ 
    BM25 & 40.6 & 38.6 & 38.2 & 37.4 & 36.6 & 36.3 & 39.5 & 37.5 & 36.8 & 36.4 & 35.5 & 37.7 & 798 & 3.7x & 1.5 & 2.7x\\
    Gzip & 63.1 & 61.0 & 59.8 & 61.1 & 60.1 & 62.3 & 57.6 & 52.9 & 51.0 & 50.1 & 50.4 & 57.2 & 824 & 3.6x & 1.5 & 2.7x \\
    SBERT & 66.9 & 61.1 & 59.0 & 61.2 & 60.3 & 64.4 & 62.6 & 56.6 & 55.1 & 53.9 & 55.0 & 59.1 & 808 & 3.6x & 1.6 & 2.5x \\
    OpenAI & 63.8 & 64.6 & 65.4 & 64.1 & 63.7 & 63.7 & 61.2 & 56.0 & 55.1 & 54.4 & 55.0 & 58.8 & 804 & 3.7x & 4.3 & 1.0x \\
    LongLLMLingua $r_k$ & 71.1 & 70.7 & 69.3 & 68.7 & 68.5 & 71.5 & 67.8 & 59.4 & 57.7 & 57.7 & 58.6 & 64.0 & 807 & 3.7x & 1.7 & 2.4x \\
    \cmidrule (r){1-1}\cmidrule (lr){2-7} \cmidrule (lr){8-13} \cmidrule (lr){14-15} \cmidrule (lr){16-17}
    \multicolumn{7}{@{}l}{{ \textit{Compression-based Methods}}} \\
    Selective-Context & 31.4 & 19.5 & 24.7 & 24.1 & 43.8 & - & 38.2 & 17.2 & 15.9 & 16.0 & 27.3 & - & 791 & 3.7x & 6.8 & 0.6x \\
    LLMLingua & 25.5 & 27.5 & 23.5 & 26.5 & 30.0 & 27.0 & 32.1 & 30.8 & 29.9 & 28.9 & 32.4 & 30.5 & 775 & 3.8x & 1.8 & 2.2x \\
    \cmidrule (r){1-1}\cmidrule (lr){2-7} \cmidrule (lr){8-13} \cmidrule (lr){14-15} \cmidrule (lr){16-17}
    {\cellcolor[rgb]{0.925,0.957,1}}\textbf{LongLLMLingua} & {\cellcolor[rgb]{0.925,0.957,1}}\textbf{75.0} & {\cellcolor[rgb]{0.925,0.957,1}}\textbf{71.8} & {\cellcolor[rgb]{0.925,0.957,1}}\textbf{71.2} & {\cellcolor[rgb]{0.925,0.957,1}}\textbf{71.2} & {\cellcolor[rgb]{0.925,0.957,1}}\textbf{74.7} & {\cellcolor[rgb]{0.925,0.957,1}}\textbf{75.5} & {\cellcolor[rgb]{0.925,0.957,1}}\textbf{68.7} & {\cellcolor[rgb]{0.925,0.957,1}}\textbf{60.5} & {\cellcolor[rgb]{0.925,0.957,1}}\textbf{59.3} & {\cellcolor[rgb]{0.925,0.957,1}}\textbf{58.3} & {\cellcolor[rgb]{0.925,0.957,1}}\textbf{61.3} & {\cellcolor[rgb]{0.925,0.957,1}}\textbf{66.7} & {\cellcolor[rgb]{0.925,0.957,1}}748 & {\cellcolor[rgb]{0.925,0.957,1}}3.9x & {\cellcolor[rgb]{0.925,0.957,1}}2.1 & {\cellcolor[rgb]{0.925,0.957,1}}2.0x \\
    \midrule
    \midrule
    Original Prompt & 75.7 & 57.3 & 54.1 & 55.4 & 63.1 & - &  68.6 & 57.4 & 55.3 & 52.5 & 55.0 & - & 2,946 & - & 4.1 & - \\
    \cmidrule (r){1-1}\cmidrule (lr){2-7} \cmidrule (lr){8-13} \cmidrule (lr){14-15} \cmidrule (lr){16-17}
    Zero-shot &  & & \multicolumn{2}{@{}c}{{ 56.1}} & & & & & \multicolumn{2}{@{}c}{{ 35.0}} & & & 15 & 196x & 1.1 & 3.7x \\
    \bottomrule
    \end{tabular}
    }
    \caption{Performance of different methods with different compression ratios (raw size / compressed size = $1/\tau$) on NaturalQuestions (20 documents)~\citep{liu2023lost}. Reorder: we reorder the documents with relevance metrics of different baselines as our document reordering strategy described in Sec. \ref{subsec:doc_reorder}. In the case of OpenAI, it corresponds to LongContextReorder$^9$ in the LangChain framework~\citep{Chase_LangChain_2022}. For results reported under 1st to 20th, we do not use the reordering strategy for all methods.
    }
    \label{tab:main_results_qa}
\end{table*}

\paragraph{Dataset \& evaluation metric}
We use NaturalQuestions for the multi-document QA task, and use LongBench and ZeroSCROLLS for general long context scenarios.
We also test on multi-hop QA tasks using MuSiQue dataset~\citep{trivedi2021musique}, and long dependency QA tasks using LooGLE benchmark~\citep{li2023loogle}. Please refer to Appendix~\ref{sec:additional_experiment} for more details on datasets.

\begin{table*}[tb]
    \small
    \centering
    \setlength{\tabcolsep}{1mm}
    \vspace{-2ex}
    \resizebox{2.0\columnwidth}{!}{
    \begin{tabular}{l|ccccccc|cc|cc}
    \toprule
        Methods & SingleDoc & MultiDoc & Summ. & FewShot & Synth. & Code & AVG & Tokens & $1/\tau$ & Latency & Speedup \\
    \midrule
    \midrule
    \multicolumn{6}{@{}r}{{ \textit{3,000 tokens constraint}}} \\
    \midrule
      \multicolumn{12}{@{}l}{{ \textit{Retrieval-based Methods}}} \\ 
    BM25 & 32.3 & 34.3 & 25.3 & 57.9 & 45.1 & 48.9 & 40.6 & 3,417 & 3x & 7.5 & 2.1x \\
    SBERT & 35.3 & 37.4 & 26.7 & 63.4 & 51.0 & 34.5 & 41.4 & 3,399 & 3x & 7.7 & 2.0x\\
    OpenAI & 34.5 & 38.6 & 26.8 & 63.4 & 49.6 & 37.6 & 41.7 & 3,421 & 3x & 13.3 & 1.2x  \\
    LongLLMLingua $r_k$ & 37.6 & 42.9 & 26.9 & 68.2 & 49.9 & 53.4 & 46.5 & 3,424 & 3x & 8.2 & 1.9x \\
    \cmidrule (r){1-1}\cmidrule (lr){2-8} \cmidrule (lr){9-10} \cmidrule (lr){11-12}
    \multicolumn{7}{@{}l}{{ \textit{Compression-based Methods}}} \\
    Selective-Context & 23.3 & 39.2 & 25.0 & 23.8 & 27.5 & 53.1 & 32.0 & 3,328 & 3x & 50.6 & 0.3x \\
    LLMLingua & 31.8 & 37.5 & 26.2 & 67.2 & 8.3 & 53.2 & 37.4 & 3,421 & 3x & 9.2 & 1.7x \\
    \cmidrule (r){1-1}\cmidrule (lr){2-8} \cmidrule (lr){9-10} \cmidrule (lr){11-12}
    {\cellcolor[rgb]{0.925,0.957,1}}\textbf{LongLLMLingua} & {\cellcolor[rgb]{0.925,0.957,1}}\textbf{40.7} & {\cellcolor[rgb]{0.925,0.957,1}}\textbf{46.2} & {\cellcolor[rgb]{0.925,0.957,1}}\textbf{27.2} & {\cellcolor[rgb]{0.925,0.957,1}}\textbf{70.6} & {\cellcolor[rgb]{0.925,0.957,1}}\textbf{53.0} & {\cellcolor[rgb]{0.925,0.957,1}}\textbf{55.2} & {\cellcolor[rgb]{0.925,0.957,1}}\textbf{48.8} & {\cellcolor[rgb]{0.925,0.957,1}}3,283 & {\cellcolor[rgb]{0.925,0.957,1}}3x & {\cellcolor[rgb]{0.925,0.957,1}}10.0 & {\cellcolor[rgb]{0.925,0.957,1}} 1.6x\\

    \midrule
    \midrule
    \multicolumn{6}{@{}r}{{ \textit{2,000 tokens constraint}}} \\
    \midrule
      \multicolumn{12}{@{}l}{{ \textit{Retrieval-based Methods}}} \\ 
    BM25 & 30.1 & 29.4 & 21.2 & 19.5 & 12.4 & 29.1 & 23.6 & 1,985 & 5x & 4.6 & 3.4x\\
    SBERT & 33.8 & 35.9 & 25.9 & 23.5 & 18.0 & 17.8 & 25.8 & 1,947 & 5x & 4.8 & 3.4x \\
    OpenAI & 34.3 & 36.3 & 24.7 & 32.4 & 26.3 & 24.8 & 29.8 & 1,991 & 5x & 10.4 & 1.5x \\
    LongLLMLingua $r_k$ & 37.8 & 41.7 & 26.9 & 66.3 & 53.0 & 52.4 & 46.3 & 1,960 & 5x & 4.7 & 3.3x\\
    \cmidrule (r){1-1}\cmidrule (lr){2-8} \cmidrule (lr){9-10} \cmidrule (lr){11-12}
    \multicolumn{7}{@{}l}{{ \textit{Compression-based Methods}}} \\
    Selective-Context & 16.2 & 34.8 & 24.4 & 15.7 & 8.4 & 49.2 & 24.8 & 1,925 & 5x & 47.1 & 0.3x \\
    LLMLingua & 22.4 & 32.1 & 24.5 & 61.2 & 10.4 & \textbf{56.8} & 34.6 & 1,950 & 5x & 5.9 & 2.6x \\
    \cmidrule (r){1-1}\cmidrule (lr){2-8} \cmidrule (lr){9-10} \cmidrule (lr){11-12}
    {\cellcolor[rgb]{0.925,0.957,1}}\textbf{LongLLMLingua} & {\cellcolor[rgb]{0.925,0.957,1}}\textbf{39.9} & {\cellcolor[rgb]{0.925,0.957,1}}\textbf{43.2} & {\cellcolor[rgb]{0.925,0.957,1}}\textbf{27.4} & {\cellcolor[rgb]{0.925,0.957,1}}\textbf{69.8} & {\cellcolor[rgb]{0.925,0.957,1}}\textbf{53.0} & {\cellcolor[rgb]{0.925,0.957,1}}56.7 & {\cellcolor[rgb]{0.925,0.957,1}}\textbf{48.3} & {\cellcolor[rgb]{0.925,0.957,1}}1,822 & {\cellcolor[rgb]{0.925,0.957,1}}6x & {\cellcolor[rgb]{0.925,0.957,1}}6.1 & {\cellcolor[rgb]{0.925,0.957,1}}2.6x\\
    \midrule
    \midrule
    Original Prompt & 39.7 & 38.7 & 26.5 & 67.0 & 37.8 & 54.2 & 44.0 & 10,295 & - & 15.6 & -\\
    \cmidrule (r){1-1}\cmidrule (lr){2-8} \cmidrule (lr){9-10} \cmidrule (lr){11-12}
    Zero-shot & 15.6 & 31.3 & 15.6 & 40.7 & 1.6 & 36.2 & 23.5 & 214 & 48x & 1.6 & 9.8x\\

    \bottomrule
    \end{tabular}
    }
    \caption{Performance of different methods under different compression ratios on LongBench~\citep{bai2023longbench} using GPT-3.5-Turbo in 2,000 tokens constraint.}
    \label{tab:main_result_long_context}
\end{table*}

\begin{table}[ht]
    \centering
	\setlength{\tabcolsep}{0.5mm}
     \resizebox{1\columnwidth}{!}{
    \begin{tabular}{lccccc}
    \toprule
         & 1st & 5th & 10th & 15th & 20th \\
         \midrule
        \textbf{LongLLMLingua} & \textbf{77.2} & \textbf{72.9} & \textbf{70.8} & \textbf{70.5} & \textbf{70.6} \\
        \multicolumn{5}{@{}l}{{ \textit{Question-aware Coarse-grained}}} \\ 
        - w/o Question-awareness & 42.1 & 40.3 & 39.7 & 40.1 & 40.3  \\
        - w/ SBERT & 73.2 & 68.5 & 65.7 & 66.1 & 66.7 \\
        - w/ $p(\mathbf{x}_{k}^{\text{doc}}|x^{\text{que}, \text{restrict}}_i)$ & 56.0 & 52.6 & 53.4 & 51.6 & 51.1 \\
        - w/o restrict & 75.1 & 72.2 & 70.3 & 70.3 & 70.2 \\
        \midrule
        - w/o Question-aware Fine-grained & 75.8 & 71.0 & 68.9 & 68.4 & 69.3 \\
        - w/o Dynamic Compression Ratio & 74.4 & 70.7 & 68.7 & 67.9 & 68.1  \\
        - w/o Subsequence Recovery & 76.7 & 71.7 & 69.4 & 69.3 & 69.7 \\
        - w/ Document Reordering & 76.2 & 76.2 & 76.2 & 76.2 & 76.2 \\
        - w/ GPT2-small & 74.6 & 71.7 & 70.1 & 69.8 & 68.5 \\ 
        \midrule
        LLMLingua & 39.7 & 39.5 & 40.4 & 37.1 & 42.3 \\
        - w/ Subsequence Recovery & 43.8 & 44.1 & 43.5 & 43.3 & 44.4\\
        \bottomrule
    \end{tabular}
    }
    \caption{Ablation study on NaturalQuestions with 2x constraint using GPT-3.5-Turbo.} 
    \label{tab:ablation}
\end{table}

\paragraph{Baselines}
We include two sets of baselines in following experiments:

\textit{(i) Retrieval-based Methods.} We assess the question-document association in the prompt using five SoTA retrieval methods: BM25, Gzip~\citep{jiang-etal-2023-low}, SentenceBERT~\citep{reimers-2019-sentence-bert}, OpenAI Embedding, and the LongLLMLingua ranker's important metric $r_k$ for coarse-grained compression. 
Notably, embedding model-based compression mirrors the method in \citet{xu2024retrieval}. 
We remove low-relevance sentences or paragraphs to meet compression limits, maintaining the original document sequence.

\textit{(ii) Compression-based Methods.} We compare our approach with two state-of-art methods for prompt compression, \ie, Selective Context~\citep{li2023unlocking} and LLMLingua~\citep{anonymous2023llmlingua}. Both methods employ LLaMA-2-7B-Chat as the small language model for compression.
In LLMLingua, a coarse-to-fine approach is used to handle constraints of compression ratio: the original prompt is first compressed to $k$ times the constraint at a coarse level, where $k$ is the granular control coefficient; token-level is then performed to reach the overall constraint. Our method follows the same coarse-to-fine logic to achieve the constraint.

\footnotetext[9]{https://python.langchain.com/docs/modules/data\_connecti on/document\_transformers/post\_retrieval/long\_context\_reorder}

\paragraph{Main results}
Table~\ref{tab:main_results_qa} and \ref{tab:main_result_long_context} present the performance of various methods under different compression constraints.
There are multiple observations and conclusions:
(1) Our \sysname{} achieves the best performance across different tasks and constraints of compression ratios.
Compared to the original prompt, our compressed prompt can derive higher performance with much lower cost.
For example, \sysname{} gains a performance boost of 21.4\% on NaturalQuestions with the ground-truth document at the 10th position, while the number of tokens input to GPT3.5-Turbo is $\sim$4x less.
(2) Compression-based methods like Selective Context~\citep{li2023unlocking} and LLMLingua~\citep{anonymous2023llmlingua} perform poorly on most tasks, especially those with abundant irrelevant information in the original prompt.
This is due to their pure information entropy based compression mechanism, which includes too much noise in the compressed results and even leads to performance worse than the zero-shot setting, \eg, on NaturalQuestions.
(3) Retrieval-based methods work well with low compression ratios. However, their performance declines as the compression progresses, \eg, $2x \rightarrow 4x$; 3000 tokens $\rightarrow$ 2000 tokens.
This may be caused by the decreased recall. Figure~\ref{sfig:recall} is the illustration of cases on NaturalQuestions.
(4) \sysname{} as well as our coarse-grained compression metric $r_k$ only is much more robust than all other baselines under different tasks and compression constraints.
With the increase of the compression ratio, \eg, $2x \rightarrow 4x$, \sysname{} even achieves a little performance gain.
We mainly owe this to the question-aware coarse-to-fine compression, which can better figure out the key information and reach a higher key information density with a higher compression ratio.
(5) The proposed reordering method helps in not only our approach but also other baselines, well demonstrating its effectiveness.
(6) Compared to the results with a 2,000 tokens constraint, overall performance of 3,000 tokens has improved. LongLLMLingua sees an increase of 1.2 points in average score and a 1.6x speedup in end-to-end latency. In this scenario, the recall rates of retrieval-based methods have increased, leading to a significant improvement in their accuracy. For example, BM25 achieves an average score of 48.9.

In addition, we also present experimental results on datasets such as MuSicQue, LooGLE, ZEROSCROLLS, etc., in Appendix~\ref{sec:additional_experiment}.

\paragraph{Ablation study}
To evaluate the contributions of different components in \sysname{},
we introduce following variants of it for ablation study.
(1) Variants about Question-aware Coarse-grained Compression, include: ours w/o Question-awareness, which calculates question-text relevance $r_k$ using information entropy in LLMLingua, ours w/ SBERT, which employs SBERT to compute $r_k$, ours w/ $p(\mathbf{x}_{k}^{\text{doc}}|x^{\text{que}, \text{restrict}}_i)$, which replace $p(x^{\text{que}, \text{restrict}}_i|\mathbf{x}_{k}^{\text{doc}})$ with $p(\mathbf{x}_{k}^{\text{doc}}|x^{\text{que}, \text{restrict}}_i)$ in Eq.~(\ref{eq:longllmlingua_coarse}), and ours w/o restrict, which only calculates the conditional probability corresponding to $x^{\text{que}}$. 
(2) Ours w/o Question-aware Fine-grained, which disregards Eq.~(\ref{eq:token_condition}) and only applies Iterative Token-level Prompt Compression as LLMLingua.
(3) Ours w/o Dynamic Compression Ratio, where all documents share the same compression ratio in fine-grained compression.
(4) Ours w/o and (5) LLMLingua w/ Subsequence Recovery, which either removes or adds the post-processing subsequence recovery strategy.
(6) Ours w/ GPT2-small, which uses the GPT2-small model as the $\mathcal{M}_S$.

Table~\ref{tab:ablation}, \ref{tab:long_bench_ablation}, and \ref{tab:musicque} shows the results of the ablation study in difference tasks.
In summary, removing any component proposed for \sysname{} will lead to a performance drop regardless of the position of the ground-truth answer.
This well validates the necessity and effectiveness of the proposed question-aware mechanism during coarse-to-fine compression, the dynamic compression ratio, and the subsequence recovery strategy.
It also shows that applying SBERT for coarse-grained compression will result in inferior performance, which implies the superiority of our question-aware importance metric in Eq.~(\ref{eq:longllmlingua_coarse}) over SBERT.
In addition, replacing $p(x^{\text{que}, \text{restrict}}_i|\mathbf{x}_{k}^{\text{doc}})$ with $p(\mathbf{x}_{k}^{\text{doc}}|x^{\text{que}, \text{restrict}}_i)$ can greatly affect performance due to the large noise in calculating $p(\mathbf{x}_{k}^{\text{doc}})$ since the perplexity of document depends on many other information besides the question. Removing the restrictive statement can increase the hallucination of small language models, leading to a decrease in performance.
Moreover, our subsequence recovery strategy can also bring performance gains for LLMLingua.
However, without our question-aware mechanism, results from LLMLingua are still less satisfactory.
For more detailed cases, please go to Appendix~\ref{sec:ablation_case}.

\paragraph{Latency evaluation}
\label{subsec:latency}

We conducte end-to-end latency testing on a V100-32G, using the prompts from Multi-document QA, LongBench, and ZeroSCROLLS in the API call, and results are shown in Table~\ref{tab:main_results_qa},~\ref{tab:main_result_long_context} and~\ref{tab:breakdowns_zero_scrolls}.
The latency includes the time cost for prompt compression and the request time for LLMs, with multiple measurements taken and averaged over. 
Results demonstrate that \sysname{} does indeed speed up the overall inference under different compression ratios and scenarios.
Moreover, with the compression ratio increasing, the acceleration effect becomes more pronounced up to 2.6x.
However, the OpenAI embedding and Selective-Context results in longer latency time, due to repeated API calls and the sequential entropy calculation of semantic units, respectively.

\section{Related Works}

\textbf{Long context for LLMs}.
Recent research has focused on expanding the %
window size of LLMs. %
Main approaches include:
(1) Staged pre-training~\citep{nijkamp2023xgen7b} which gradually increases the context window;
(2) Modifying~\citep{press2022train} or interpolating position embeddings~\citep{chen2023extending,peng2023yarn};
(3) Using linear or sparse attention mechanisms~\citep{ding2023longnet, sun2023retentive};
(4) Utilizing external memory modules for context storage~\citep{bertsch2023unlimiformer,tworkowski2023focused}.
While these methods address context window expansion, their impact on downstream task performance has yet to be discussed.

\textbf{Information distribution in prompt}. 
Recent empirical experiments have shown that LLM performance decreases with less effective information in a prompt~\citep{bai2023longbench,longchat2023,shi2023large}.
Moreover, the position of relevant information in a prompt has a significant impact on performance~\citep{wu2023selfadaptive}. \citet{liu2023lost} suggests that LLMs have more difficulty comprehending information located in the middle of a prompt compared to those at the edges. 

\textbf{Retrieval methods} can be categorized as dense or sparse retrieval methods. Sparse retrieval methods, like BM25, determine the relevance between queries and documents based on n-gram information. Conversely, dense retrieval methods assess the relevance between queries and documents in latent space using embedding model~\citep{reimers-2019-sentence-bert, bge_embedding, günther2023jina} and reranker model~\citep{bge_embedding}. Recently, \citet{jiang-etal-2023-low} proposed an unsupervised dense retrieval method that leverages traditional compression algorithms, such as gzip, and k-nearest neighbors.

\textbf{Prompt compression methods} can be grouped into three main categories: (1) Token pruning~\citep{goyal2020power, kim2021length, modarressi2022adapler} %
and token merging~\citep{bolya2023token}, which need model fine-tuning or intermediate results during inference and have been used with BERT-scale models. %
(2) Soft prompt tuning methods like GIST~\citep{mu2023learning}, AutoCompressor~\citep{chevalier2023adapting}, and ICAE~\citep{ge2023incontext}, which require LLMs' %
parameter fine-tuning, %
making them suitable for specific domains but not directly applicable to black-box LLMs.
(3) Information-entropy-based approaches such as Selective Context~\citep{li2023unlocking} and LLMLingua~\citep{anonymous2023llmlingua}, which use a small language model to calculate the self-information or perplexity of each token in the original prompt and then remove tokens with lower perplexities.

\section{Conclusion}
We propose \sysname{} to address the three challenges, \ie,  higher computational cost, performance reduction, and position bias for LLMs in long context scenarios.
We develop \sysname{} from the perspective of efficient prompt compression, thus reducing computational cost.
We further design four components, \ie, a question-aware coarse-to-fine compression method, a document reordering mechanism, dynamic compression ratios, and a subsequence recovery strategy to improve LLMs’ perception of the key information, with which \sysname{} demonstrate superior performance.
Experiments on the multi-document QA, multi-hop QA, and long context benchmarks demonstrate that \sysname{} 
compressed prompt can derive higher performance than original prompts while both API costs for inference and the end-to-end system latency are largely reduced.

\section*{Limitation}

Although previous experiments demonstrate \sysname{}'s effectiveness and efficiency across a broad range of tasks, the method still has the following limitations:
1) LongLLMLingua is a question-aware approach, meaning it requires re-compression for different questions, even with the same context, preventing caching of the context. Moreover, in terms of computational cost, LongLLMLingua increases the computation by twice as much as LLMLingua. This can lead to greater overhead in real-world applications. However, this issue can be mitigated by extending the question-aware approach to a task-aware approach, allowing for reuse and caching.
2) While the effectiveness of LongLLMLingua has been tested on a wide range of tasks, especially on the multi-hop QA dataset MuSicQue~\citep{trivedi2021musique}, its effectiveness might be impacted when the relationship between context and prompt is more complex and subtle due to the coarse-level question-aware approach.

\bibliography{acl2024}

\appendix

\section{Derivation Of Question-Aware Fine-Grained Compression}
\label{sec:derivation}

Based on the definition of Eq.~(\ref{eq:token_condition}), we can derive that,
\begin{equation}
     \begin{aligned}
         s_i &= \text{perplexity}(x_i|x_{<i}) - \text{perplexity}(x_i|x^{\text{que}}, x_{<i}) \\
         &= q(x_i)\log p(x_i|x^{\text{que}}, x_{<i}) - q(x_i)\log p(x_i|x_{<i}) \\
         &= q(x_i)\log\frac{p(x_i|x^{\text{que}}, x_{<i})}{p(x_i|x_{<i})} \\
     \end{aligned}
\end{equation}
In the actual calculation of perplexity, a log operation is performed to avoid overflow, and $q(x_i)$ represents the probability distribution of the ground-truth.

At the same time, we can derive the following expanded expression based on Bayes' theorem.
\begin{equation}
\begin{aligned}
    p(x^{\text{que}}|x_i, x_{<i}) &= \frac{p(x_i|x^{\text{que}}, x_{<i})p(x^{\text{que}})}{p(x_i|x_{<i})} \\&= p(x^{\text{que}})  \frac{p(x_i|x^{\text{que}}, x_{<i})}{p(x_i|x_{<i})}
    \label{eq:bayes}
\end{aligned}
\end{equation}

The probability distribution $p(x^\text{que})$ of the question and the ground-truth distribution $q(x_i)$ of $x_i$  are constants, hence $s_i$ can be considered as the representation of Eq.~(\ref{eq:bayes}).
\begin{equation}
    s_i \propto p(x^{\text{que}}|x_i, x_{<i})
\end{equation}

So we can utilize Eq.~(\ref{eq:token_condition}) to represent the probability distribution $p(x^{\text{que}}|x_i, x_{<i})$, which represents the condition likelihood of generating $x^\text{que}$ given the token $x_i$.
Therefore, we can represent the token-level sensitive distribution for the question $x^{\text{que}}$ using just a single inference.
For tokens that are unrelated to $x^{\text{que}}$, such as the tokens on the right side of Figure~\ref{sfig:ppl}, their original amount of information may be high, but the contrastive perplexity remains at a relatively low level.
Finally, we observe that the form of contrastive perplexity is equivalent to conditional pointwise mutual information~\citep{church1990word}.

\section{Experiment Details}
\label{sec:experiment_details}

\subsection{Dataset Details}

We use NaturalQuestions~\citep{liu2023lost} for the multi-document QA task, MuSicQue~\citep{trivedi2021musique} for the multi-hop QA task, and use LongBench~\citep{bai2023longbench}, ZeroSCROLLS~\citep{shaham2023zeroscrolls}, LooGLE~\citep{li2023loogle} for general long context scenarios.
The specific details of the dataset are as follows:

\paragraph{NaturalQuestions multi-document QA}
A multi-document question-answering dataset, comprising 2,655 problems, was built by \cite{liu2023lost} based on the NaturalQuestions dataset~\citep{kwiatkowski2019natural}. This dataset provides a realistic retrieval-augmented generation setup that closely resembles commercial search and question-answering applications (e.g., Bing Chat). Each example in the dataset contains a question and k related documents, utilizing the Contriever retrieval system~\citep{izacard2022unsupervised}, one of which includes a document with the correct answer. To perform this task, the model must access the document containing the answer within its input context and use it to answer the question. The dataset's data is sourced from the NaturalQuestions dataset, which contains historical queries issued to the Google search engine and human-annotated answers extracted from Wikipedia. The average prompt token length in this benchmark is 2,946. For our experiments, we used the version provided by \cite{liu2023lost} that includes 20 documents\footnote[10]{https://github.com/nelson-liu/lost-in-the-middle}.
The dataset comprises five different ground truth document position settings in the prompt: 1st, 5th, 10th, 15th, and 20th.

\paragraph{LongBench}
A multi-task long context benchmark consists of 3,750 problems in English and includes six categories with a total of 16 tasks. These tasks encompass key long-text application scenarios, such as single-document QA, multi-document QA, summarization, few-shot learning, synthetic tasks, and code completion. The average prompt token length in this benchmark is 10,289. For our experiments, we used the English dataset and evaluation scripts provided by \cite{bai2023longbench} for this benchmark\footnote[11]{https://github.com/THUDM/LongBench}.

\paragraph{ZeroSCROLLS}
The multi-task long context benchmark consists of 4,378 problems, including four categories with a total of 10 tasks. These tasks cover summarization, question answering, aggregated sentiment classification, and information reordering. The average prompt token length in this benchmark is 9,788. For our experiments, we used the validation set and evaluation scripts provided by \cite{shaham2023zeroscrolls} for this dataset\footnote[12]{https://www.zero.scrolls-benchmark.com/}.

\paragraph{MuSiQue}
The multi-hop question-answer dataset is composed of 39,876, 4,834, and 4,918 problems in the training, validation, and testing datasets, respectively. This dataset requires the language model to conduct multiple inferences based on the content of several documents and provide corresponding answers, thereby necessitating a certain capability for global information processing. The average token length for prompts in this dataset is 2,477. For our experiments, we utilized the validation set and evaluation scripts provided by \cite{trivedi2021musique} for this dataset\footnote[13]{https://github.com/stonybrooknlp/musique}.

\paragraph{LooGLE}
The multi-task long context benchmark comprises 6,448 problems, divided into three categories: summarization, short dependency question answering, and long dependency question answering. The average prompt token length in this benchmark stands at 24,005. For our experiments, we focused on the long dependency question answering subset, which includes four types of tasks: information retrieval, timeline reordering, computation, and comprehension. This subset contains 1,101 problems. We utilized the evaluation scripts provided by \cite{li2023loogle} for this dataset\footnote[14]{https://github.com/bigai-nlco/LooGLE}.

\subsection{Other Implementation Details}
\label{sec:appendix_restrict}
All experiments were conducted using a Tesla V100 (32GB).
We use tiktoken\footnote[15]{https://github.com/openai/tiktoken} and GPT-3.5-Turbo model to count all the tokens.
We set the granular control coefficient $k$ to $2$.
We use the pre-defined compression rates $\tau_{\text{ins}}=0.85$ and $\tau_{\text{que}}=0.9$ for instructions and questions.
The segment size used in the iterative token-level compression is set to $200$.
The $\delta\tau$ used in dynamic compression ratio is set to 0.3.
For a fair comparison, we only used reordering in the NaturalQuestions Multi-document QA and noted this in Table~\ref{tab:main_results_qa}.
We use ``\textit{We can get the answer to this question in the given documents.}" as the guideline sentence in Eq.~(\ref{eq:token_condition}).

\begin{figure*}[htb]
  \centering
  \subfloat[1st]{
    \label{sfig:ppl_doc_0}
    \includegraphics[height=0.48\columnwidth]{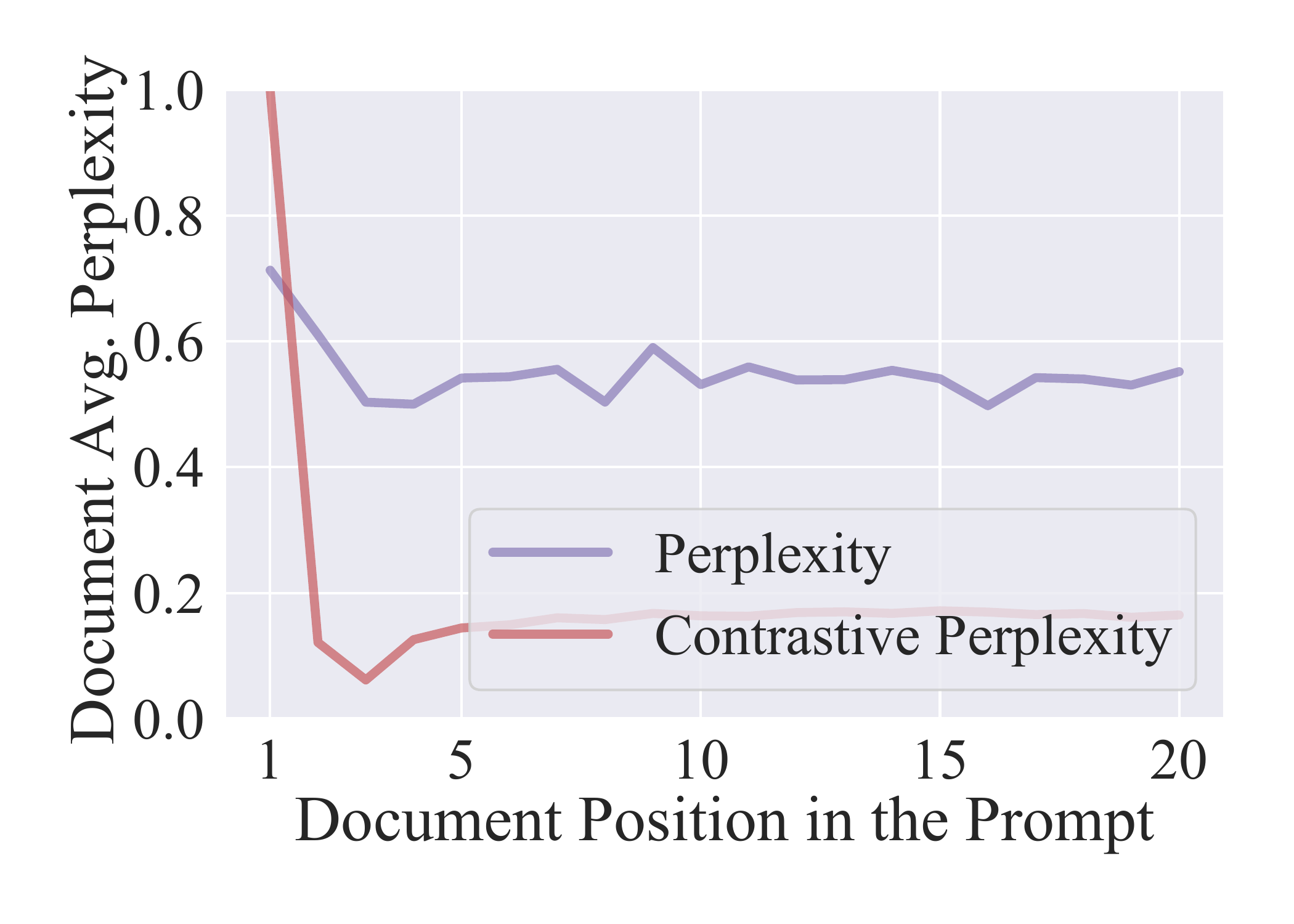}}
  \subfloat[10th]{
    \label{sfig:ppl_doc_9}
    \includegraphics[height=0.48\columnwidth]{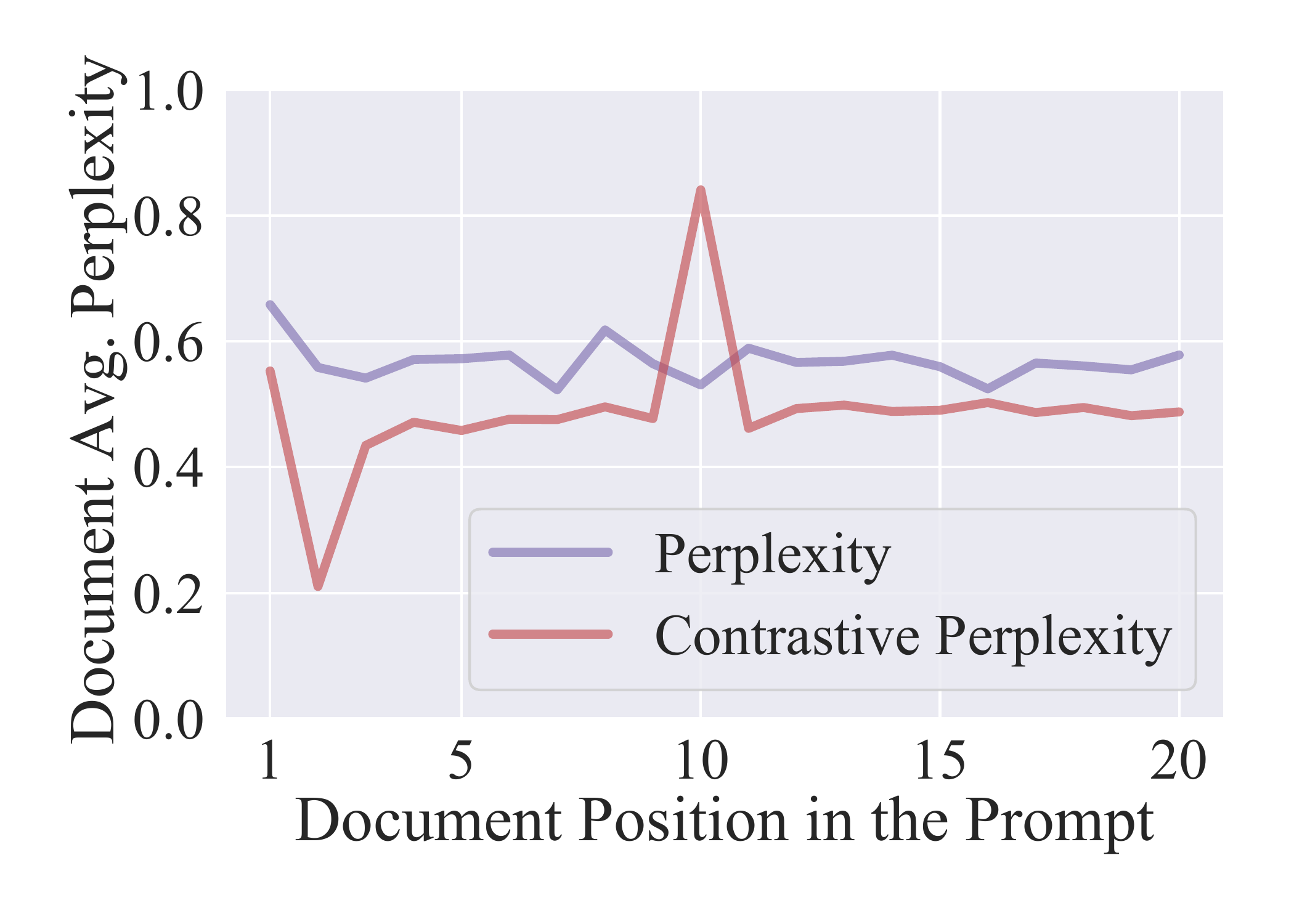}}
  \subfloat[15th]{
    \label{sfig:ppl_doc_14}
    \includegraphics[height=0.48\columnwidth]{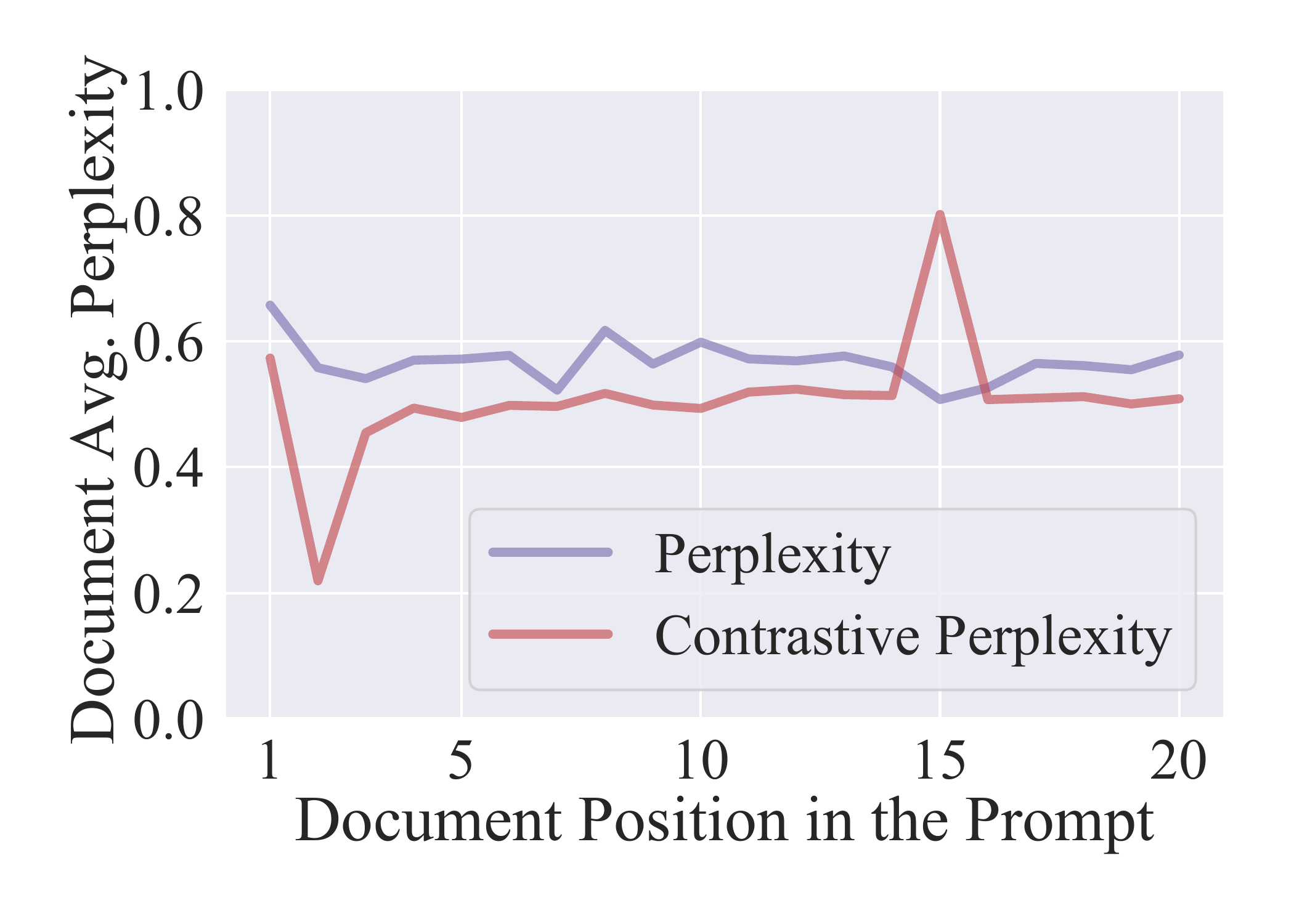}}
  \caption{The distribution of document-level average perplexity when the ground-truth document is in different positions.}
  \label{fig:doc_ppl}
\end{figure*}
\begin{table*}[htb]
    \small
    \centering
    \setlength{\tabcolsep}{1mm}
    \vspace{-2ex}
    \resizebox{1.9\columnwidth}{!}{
    \begin{tabular}{l|ccccccccc}
    \toprule
        Methods & SingleDoc & MultiDoc & Summ. & FewShot & Synth. & Code & AVG & Tokens & $1/\tau$ \\
         \cmidrule (r){1-1}\cmidrule (lr){2-10}
        \textbf{LongLLMLingua} & 39.9 & 43.2 & 27.4 & 69.8 & 53.0 & 56.7 & 48.3 & 1,822 & 6x \\
        \cmidrule (r){1-1}\cmidrule (lr){2-10}
        \multicolumn{5}{@{}l}{{ \textit{Question-aware Coarse-grained}}} \\ 
        - w/o Question-awareness & 27.1 & 38.7 & 25.4 & 62.0 & 18.0 & 53.3 & 37.4 & 1,945 & 5x \\
        - w/ SBERT & 34.0 & 38.7 & 24.1 & 57.9 & 32.5 & 31.1 & 36.4 & 1,790 & 6x \\
        - {w/ $p(\mathbf{x}_{k}^{\text{doc}}|x^{\text{que}, \text{restrict}}_i)$} & 22.5 & 28.9 & 23.2 & 53.0 & 22.5 & 33.3 & 30.6 & 1,794 & 6x \\
        - {w/o restrict} & 37.8 & 39.5 & 26.4 & 64.8 & 52.5 & 55.8 & 46.1 & 1,834 & 6x \\
        \cmidrule (r){1-1}\cmidrule (lr){2-10}
        - w/o Question-aware Fine-grained & 35.7 & 41.1 & 26.4 & 62.9 & 44.5 & 54.8 & 44.2 & 1,807 & 6x \\
        - w/o Dynamic Compression Ratio & 36.1 & 40.6 & 26.9 & 67.2 & 48.0 & 55.8 & 45.7 & 1,851 & 6x  \\
        - w/o Subsequence Recovery & 38.6 & 41.8 & 27.3 & 69.0 & 53.8 & 56.6 & 47.8 & 1,809 & 6x \\
        - w/o Document Reordering & 39.0 & 42.2 & 27.4 & 69.3 & 53.8 & 56.6 & 48.0 & 1,809 & 6x \\
        - {w/ GPT2-small} & 35.9 & 39.4 & 25.0 & 60.6 & 42.0 & 55.4 & 43.0 & 1,892 & 5x \\ 
    \bottomrule
    \end{tabular}
    }
    \caption{Ablation on LongBench~\citep{bai2023longbench} using GPT-3.5-Turbo in 2,000 tokens constraint.}
    \label{tab:long_bench_ablation}
\end{table*}

For the baselines experiment, we use the currently recommended strongest model, all-mpnet-base-v2\footnote[16]{https://www.sbert.net/docs/pretrained\_models.html}, as the dense representation model for SentenceBERT. 
We use the recommended ``text-embedding-ada-002" as the embedding model for OpenAI Embedding\footnote[17]{https://platform.openai.com/docs/guides/embeddings/}.
We use the GPT2-dolly\footnote[18]{https://huggingface.co/lgaalves/gpt2-dolly} as the small language model in w/ GPT2-small ablation experiments.

\section{Additional Experimental Results}
\label{sec:additional_experiment}

\subsection{Empirical Study of Question-aware Fine-grained Compression}
\label{subsec:empirical}

Figure~\ref{fig:doc_ppl} shows the distribution of the document's average perplexity when the ground-truth is located at more positions within the prompt. As can be observed, as the context length increases, the original perplexity curve remains relatively stable. In unrelated documents, a higher perplexity is still retained, making it easier to remove relevant tokens from the related documents in the prompt compression process, thereby damaging the corresponding semantic information. Contrarily, contrastive perplexity shows an increase in perplexity in documents related to the question. According to the theoretical derivation in Appendix~\ref{sec:derivation}, it's known that contrastive perplexity characterizes the conditional probability of tokens corresponding to the question. The higher the relevance, the higher the contrastive perplexity, thereby retaining key information in the prompt compression process.

\subsection{Ablation in LongBench}

\begin{table*}[htb]
    \small
    \centering
    \setlength{\tabcolsep}{1mm}
    \vspace{-2ex}
    \resizebox{1.75\columnwidth}{!}{
    \begin{tabular}{l|ccccccccc}
    \toprule
        Methods & SingleDoc & MultiDoc & Summ. & FewShot & Synth. & Code & AVG & Tokens & $1/\tau$ \\
         \cmidrule (r){1-1}\cmidrule (lr){2-10}
    Original Prompt & 27.4 & 30.3 & 20.3 & 49.9 & 12.5 & 42.5 & 30.5 & 10,295 & -\\
    \cmidrule (r){1-1}\cmidrule (lr){2-10}
      \multicolumn{10}{@{}l}{{ \textit{Retrieval-based Methods}}} \\ 
    BM25 & 2.4 & 2.6 & 16.4 & 8.7 & 0.0 & 44.7 & 12.5 & 1,985 & 5x \\
    SBERT & 11.6 & 13.7 & 21.1 & 16.2 & 7.5 & 30.0 & 16.7 & 1,947 & 5x\\
    LongLLMLingua $r_k$ & 30.3 & 32.4 & 24.5 & 41.0 & 27.5 & 38.1 & 32.3 & 1,960 & 5x \\
    \cmidrule (r){1-1}\cmidrule (lr){2-10}
    \multicolumn{7}{@{}l}{{ \textit{Compression-based Methods}}} \\
    Selective-Context & 16.1 & 23.5 & 21.8 & 21.4 & 2.5 & 35.9 & 20.2 & 1,925 & 5x \\
    LLMLingua & 20.6 & 22.3 & 22.4 & 35.6 & 0.0 & 35.4 & 22.7 & 1,950 & 5x \\
    \cmidrule (r){1-1}\cmidrule (lr){2-10}
    {\cellcolor[rgb]{0.925,0.957,1}}\textbf{LongLLMLingua} & {\cellcolor[rgb]{0.925,0.957,1}}31.3 & {\cellcolor[rgb]{0.925,0.957,1}}34.6 & {\cellcolor[rgb]{0.925,0.957,1}}24.6 & {\cellcolor[rgb]{0.925,0.957,1}}46.1 & {\cellcolor[rgb]{0.925,0.957,1}}27.8 & {\cellcolor[rgb]{0.925,0.957,1}}48.8 & {\cellcolor[rgb]{0.925,0.957,1}}35.5 & {\cellcolor[rgb]{0.925,0.957,1}}1,822 & {\cellcolor[rgb]{0.925,0.957,1}}6x \\
    \bottomrule
    \end{tabular}
    }
    \caption{Performance of different methods under different compression ratios on LongBench~\citep{bai2023longbench} using LongChat-13b in 2,000 tokens constraint. 
    }
    \label{tab:long_bench_longchat}
\end{table*}

\begin{table*}[htb]
    \small
    \centering
    \setlength{\tabcolsep}{1mm}
    \vspace{-2ex}
    \resizebox{2.05\columnwidth}{!}{
    \begin{tabular}{l|ccccccccccc|cc|cc}
    \toprule
        Methods &  GvRp & SSFD & QMsm & SQAL & QALT & Nrtv & Qspr & MuSQ & SpDg & BkSS & AVG & Tokens & $1/\tau$ & Latency & Speedup \\
    \midrule
    \midrule    
    \multicolumn{16}{@{}c}{{ \textit{3,000 tokens constraint}}} \\
    \midrule
      \multicolumn{10}{@{}l}{{ \textit{Retrieval-based Methods}}} \\ 
    BM25 & 9.7 & 3.4 & 11.7 & 14.3 & 57.1 & 5.9 & 25.7 & 11.2 & 29.6 & 29.6 & 19.8 & 3,379 & 3x & 5.5 & 2.2x \\
    SBERT & 16.5 & 9.8 & 12.3 & 15.2 & 60.0 & 14.6 & 23.4 & 12.1 & 39.4 & 36.4 & 24.0 & 3,340 & 3x & 5.9 & 2.1x \\
    OpenAI & 14.3 & 8.3 & 12.0 & 15.3 & 66.7 & 13.3 & 24.3 & 11.7 & 31.2 & 26.4 & 22.4 & 3,362 & 3x & 11.7 & 1.0x  \\
    LongLLMLingua $r_k$ & 19.5 & 11.6 & 14.7 & 15.5 & 66.7 & 20.5 & 27.6 & 13.0 & 60.8 & 43.4 & 29.3 & 3,350 & 3x & 6.2 & 2.0x  \\
    \cmidrule (r){1-1} \cmidrule (lr){2-12} \cmidrule (lr){13-14} \cmidrule (lr){15-16}
    \multicolumn{7}{@{}l}{{ \textit{Compression-based Methods}}} \\
    Selective-Context & 20.8 & 9.1 & 11.7 & 13.4 & 50.0 & 9.8 & 26.1 & 11.0 & 46.0 & 9.5 & 20.7 & 3,460 & 3x & 54.2 & 0.2x  \\
    LLMLingua & 18.7 & 10.0 & 14.9 & 16.8 & 61.9 & 26.9 & 27.2 & 23.4 & 62.9 & 44.5 & 30.7 & 3,366 & 3x & 7.4 & 1.7x  \\
    \cmidrule (r){1-1} \cmidrule (lr){2-12} \cmidrule (lr){13-14} \cmidrule (lr){15-16}
    {\cellcolor[rgb]{0.925,0.957,1}}\textbf{LongLLMLingua} & {\cellcolor[rgb]{0.925,0.957,1}}22.1 & {\cellcolor[rgb]{0.925,0.957,1}}12.8 & {\cellcolor[rgb]{0.925,0.957,1}}15.9 & {\cellcolor[rgb]{0.925,0.957,1}}17.1 & {\cellcolor[rgb]{0.925,0.957,1}}67.0 & {\cellcolor[rgb]{0.925,0.957,1}}27.8 & {\cellcolor[rgb]{0.925,0.957,1}}31.3 & {\cellcolor[rgb]{0.925,0.957,1}}23.9 & {\cellcolor[rgb]{0.925,0.957,1}}65.8 & {\cellcolor[rgb]{0.925,0.957,1}}46.5 & {\cellcolor[rgb]{0.925,0.957,1}}33.0 & {\cellcolor[rgb]{0.925,0.957,1}}3,431 & {\cellcolor[rgb]{0.925,0.957,1}}3x & {\cellcolor[rgb]{0.925,0.957,1}}8.2 & {\cellcolor[rgb]{0.925,0.957,1}}1.5x \\
    \midrule
    \midrule
    \multicolumn{16}{@{}c}{{ \textit{2,000 tokens constraint}}} \\
    \midrule
      \multicolumn{10}{@{}l}{{ \textit{Retrieval-based Methods}}} \\ 
    BM25 & 8.8 & 2.5 & 11.1 & 13.5 & 60.0 & 7.0 & 4.9 & 20.3 & 39.9 & 32.9 & 20.1 & 1,799 & 5x & 3.8 & 3.2x  \\
    SBERT & 10.2 & 7.9 & 13.7 & 13.2 & 60.0 & 8.1 & 10.8 & 1.7 & 37.2 & 42.8 & 20.5 & 1,773 & 6x & 4.1 & 3.0x  \\
    OpenAI & 11.1 & 8.0 & 11.8 & 13.6 & 60.0 & 7.1 & 13.2 & 4.0 & 33.6 & 43.6 & 20.6 & 1,784 & 5x & 9.9 & 1.2x \\
    LongLLMLingua $r_k$ & 18.2 & 9.8 & 12.3 & 15.9 & 57.1 & 10.1 & 17.8 & 7.3 & 57.7 & 42.3 & 24.9 & 1,771 & 6x & 4.7 & 2.6x  \\
    \cmidrule (r){1-1} \cmidrule (lr){2-12} \cmidrule (lr){13-14} \cmidrule (lr){15-16}
    \multicolumn{7}{@{}l}{{ \textit{Compression-based Methods}}} \\
    Selective-Context & 19.0 & 8.4 & 9.7 & 12.4 & 47.0 & 12.5 & 21.6 & 11.5 & 41.2 & 11.0 & 19.4 & 1,865 & 5x & 47.5 & 0.3x  \\
    LLMLingua & 19.4 & 11.9 & 13.1 & 16.0 & 62.1 & 23.7 & 24.0 & 22.4 & 33.9 & 44.9 & 27.2 & 1,862 & 5x & 4.8 & 0.3x  \\
    \cmidrule (r){1-1} \cmidrule (lr){2-12} \cmidrule (lr){13-14} \cmidrule (lr){15-16}
    {\cellcolor[rgb]{0.925,0.957,1}}\textbf{LongLLMLingua} & {\cellcolor[rgb]{0.925,0.957,1}}20.1 & {\cellcolor[rgb]{0.925,0.957,1}}12.4 & {\cellcolor[rgb]{0.925,0.957,1}}14.9 & {\cellcolor[rgb]{0.925,0.957,1}}16.5 & {\cellcolor[rgb]{0.925,0.957,1}}65.1 & {\cellcolor[rgb]{0.925,0.957,1}}27.7 & {\cellcolor[rgb]{0.925,0.957,1}}30.7 & {\cellcolor[rgb]{0.925,0.957,1}}23.6 & {\cellcolor[rgb]{0.925,0.957,1}}68.5 & {\cellcolor[rgb]{0.925,0.957,1}}47.2 & {\cellcolor[rgb]{0.925,0.957,1}}32.7 & {\cellcolor[rgb]{0.925,0.957,1}}1,826 & {\cellcolor[rgb]{0.925,0.957,1}}6x & {\cellcolor[rgb]{0.925,0.957,1}}5.2 & {\cellcolor[rgb]{0.925,0.957,1}}2.3x \\
    \midrule
    \midrule
    Original Prompt & 21.8 & 12.1 & 17.9 & 17.4 & 66.7 & 25.3 & 29.8 & 20.0 & 69.7 & 44.1 & 32.5 & 9,788 & - & 12.2 & - \\
    \cmidrule (r){1-1} \cmidrule (lr){2-12} \cmidrule (lr){13-14} \cmidrule (lr){15-16}
    Zero-shot & 9.4 & 3.0 & 8.6 & 11.4 & 42.9 & 10.6 & 12.4 & 5.5 & 4.2 & 0.0 & 12.8 & 32 & 306x & 1.0 & 12.2x \\
    \bottomrule
    \end{tabular}
    }
    \caption{Performance breakdown of different methods under different compression ratios on ZeroSCROLLS~\citep{shaham2023zeroscrolls} using GPT-3.5-Turbo.}
    \label{tab:breakdowns_zero_scrolls}
\end{table*}

\begin{table}[!htb]
    \centering
	\setlength{\tabcolsep}{0.5mm}
     \resizebox{0.9\columnwidth}{!}{
    \begin{tabular}{lccc}
    \toprule
         Methods & F1 & Tokens & $1/\tau$ \\
         \midrule
        Original Prompt & 45.8 & 2,427 & - \\
        BM25 & 28.5 & 1,295 & 1.9x \\
        SBERT & 36.2 & 1,288 & 1.9x\\
        LongLLMLingua $r_k$ & 46.3 & 1,295 & 1.9x \\
        Selective-Context & 19.6 & 1,141 & 2.1x \\
        LLMLingua & 40.1 & 1,110 & 2.2x \\
        \cmidrule (r){1-1}\cmidrule (lr){2-4}
        \textbf{LongLLMLingua} & \textbf{51.2} & 1,077 & 2.3x \\
        \multicolumn{4}{@{}l}{{ \textit{Question-aware Coarse-grained}}} \\ 
        - w/o Question-awareness & 43.2 & 1,076 & 2.3x \\
        - w/ SBERT & 47.3 & 1,070 & 2.3x \\
        - {w/ $p(\mathbf{x}_{k}^{\text{doc}}|x^{\text{que}, \text{restrict}}_i)$} & 44.0 & 1,066 & 2.3x \\
        - {w/o restrict} & 49.2 & 1,078 & 2.3x \\
        \midrule
        - w/o Question-aware Fine-grained & 48.4 & 1,118 & 2.2x \\
        - w/o Dynamic Compression Ratio & 48.2 & 1,090 & 2.2x \\
        - w/o Subsequence Recovery & 50.7 & 1,077 & 2.3x \\
        - w/o Document Reordering & 49.2 & 1,077 & 2.3x \\
        - {w/ GPT2-small} & 48.4 & 1,095 & 2.2x\\ 
        \bottomrule
    \end{tabular}
    }
    \caption{Performance of different methods and ablation study on MuSicQue~\citep{trivedi2021musique} with 2x constraint using GPT-3.5-Turbo.}
    \label{tab:musicque}
\end{table}

\begin{table*}[htb]
    \small
    \centering
    \setlength{\tabcolsep}{1mm}
    \vspace{-2ex}
    \resizebox{1.9\columnwidth}{!}{
    \begin{tabular}{l|ccccccc}
    \toprule
        Methods & Retrieval & Timeline Reorder & Computation & Reasoning & AVG & Tokens & $1/\tau$ \\
    \cmidrule (r){1-1}\cmidrule (lr){2-8}
      \multicolumn{8}{@{}l}{{ \textit{Retrieval-based Methods}}} \\ 
    BM25 & 20.4 & 21.7 & 8.2 & 26.3 & 19.2 & 3,185 & 10x\\
    SBERT & 28.9 & 21.1 & 10.7 & 27.2 & 22.0 & 3,169 & 10x \\
    LongLLMLingua $r_k$ & 38.6 & 32.2 & 16.2 & 26.3 & 28.3 & 3,158 & 10x \\
    \cmidrule (r){1-1}\cmidrule (lr){2-8}
    \multicolumn{8}{@{}l}{{ \textit{Compression-based Methods}}} \\
    Selective-Context & 16.7 & 5.0 & 2.3 & 17.6 & 10.4 & 3,710 & 8x \\
    LLMLingua & 10.0 & 25.0 & 13.3 & 21.1 & 17.3 & 3,404 & 9x \\
    \cmidrule (r){1-1}\cmidrule (lr){2-8}
    {\cellcolor[rgb]{0.925,0.957,1}}\textbf{LongLLMLingua} & {\cellcolor[rgb]{0.925,0.957,1}}\textbf{40.0} & {\cellcolor[rgb]{0.925,0.957,1}}\textbf{35.0} & {\cellcolor[rgb]{0.925,0.957,1}}\textbf{19.7} & {\cellcolor[rgb]{0.925,0.957,1}}\textbf{33.6} & {\cellcolor[rgb]{0.925,0.957,1}}\textbf{32.1} & {\cellcolor[rgb]{0.925,0.957,1}}3,121 & {\cellcolor[rgb]{0.925,0.957,1}}10x \\
    {\cellcolor[rgb]{0.925,0.957,1}}\textbf{LongLLMLingua} w/o Reorder & {\cellcolor[rgb]{0.925,0.957,1}}39.3 & {\cellcolor[rgb]{0.925,0.957,1}}33.8 & {\cellcolor[rgb]{0.925,0.957,1}}18.7 & {\cellcolor[rgb]{0.925,0.957,1}}31.6 & {\cellcolor[rgb]{0.925,0.957,1}}30.9 & {\cellcolor[rgb]{0.925,0.957,1}}3,119 & {\cellcolor[rgb]{0.925,0.957,1}}10x \\
    \midrule
    Original Prompt & 24.1 & 20.9 & 13.5 & 32.1 & 22.6 & 30,546 & - \\
    \cmidrule (r){1-1}\cmidrule (lr){2-8}
    Zero-shot & 8.7 & 6.3 & 1.2 & 14.5 & 7.7 & 43 & 710x  \\
    \bottomrule
    \end{tabular}
    }
    \caption{Performance of different methods on LooGLE~\citep{li2023loogle} long dependency QA.}
    \label{tab:result_loogle}
\end{table*}

Table~\ref{tab:long_bench_ablation} presents the results from the ablation experiment in the LongBench long context benchmark. It can be observed that in various long context tasks: 1) Removing the question-aware coarse-grained, question-aware fine-grained, dynamic compression ratio, document reordering, and subsequence recovery proposed by LongLLMLingua all result in different degrees of performance drop. 2) Among these, question-aware coarse-grained is particularly important for document-based QA and synthetic tasks, with the maximum drop being 35.8 points; its impact on summarization and code tasks is relatively smaller. 3) The design of the conditional probability in the question-aware coarse-grained module improves the results in all tasks, including code completion, single-document question-answer, and synthetic tasks. Changing the order of conditional probabilities or removing the restrict prompt both lead to varying degrees of performance decline. 4) Removing question-aware fine-grained, dynamic compression ratio has a more significant impact on document-based QA and synthetic tasks. 5) The subsequence recovery module can enhance reference-based tasks, but its improvement on tasks like summarization, code, synthetic, etc., is relatively smaller. 6) Document reordering is effective for all types of tasks. Reordering at the document level does not affect LLMs' understanding of context information, even for timeline-related tasks (see timeline reorder in LooGLE, Table~\ref{tab:result_loogle}). On the contrary, reordering can effectively alleviate the "lost in the middle" issue, thereby improving LLMs performance. 7) Using GPT2-small reduces the capture of effective tokens, but it can still achieve results close to or even slightly better than the original prompt.

\subsection{LongBench Using LongChat-13b-16k}

Table~\ref{tab:long_bench_longchat} presents the experiment results in the LongBench long context benchmark using LongChat-13b-16k. It can be seen that the compressed prompt can also achieve good results on other LLMs, such as LongChat-13b-16k. Specifically, 1) there is a maximum improvement of 15.5 points in synthetic tasks. Except for a slight drop in few-shot Learning, there is an improvement of 3-5 points in other tasks. 2) The performance trends of retrieval-based and compressed-based baselines are similar to the results in GPT-3.5-Turbo.

\subsection{ZeroSCROLLS}
\begin{table*}[htb]
    \small
    \centering
    \setlength{\tabcolsep}{2mm}
    \vspace{-2ex}
     \resizebox{1.8\columnwidth}{!}{
    \begin{tabular}{lccccc}
    \toprule
         & Multi-document QA & LongBench & ZeroScolls & MuSicQue & LooGLE\\
         \midrule
        Original & 4.6 & 31.5 & 30.6 & 3.8 & 93.6  \\
        Ours & 1.3 ($\downarrow$71.7\%) & 3.0 ($\downarrow$90.5\%) & 3.2 ($\downarrow$89.5\%) & 1.8 ($\downarrow$52.6\%) & 5.6 ($\downarrow$94.0\%) \\
        \bottomrule
    \end{tabular}
    }
    \caption{The inference costs \$ (per 1,000 samples) for various datasets using GPT-3.5-Turbo.}
    \label{tab:cost}
\end{table*}

Table~\ref{tab:breakdowns_zero_scrolls} presents a detailed performance breakdown on the ZeroSCROLLS benchmark. It can be observed that in the four summarization tasks - GvRp, SSFD, QMsm, SQAL, LongLLMLingua closely matches or slightly surpasses the original results under two compression constraints. Meanwhile, in the four long context QA tasks - Qsqr, Nrtv, QALT, MuSQ, there is a significant improvement. Notably, in the MuSiQue task, which is based on a question-answering dataset from books and movie scripts, there is a 2.1 point increase even under a 2,000 tokens constraint. It's worth mentioning that MuSiQue is a multi-hop question-answering dataset that requires LLMs to utilize global information for long dependency QA. LongLLMLingua can also improve by 3.5 points under a 6x compression ratio. In the two ordering tasks, SpDg and BkSS, LongLLMLingua can better retain globally sensitive information, resulting in a 3.0 point improvement in BkSS after prompt compression.

It's important to note that although the ZeroScrolls validation dataset is relatively small, it still demonstrates conclusions similar to previous experimental observations across various methods and tasks. Furthermore, this study conducted an in-depth analysis of the multi-hop QA task - MuSiQue, and another long context benchmark - LooGLE. The results can be found in Appendix~\ref{sec:musq} and Appendix~\ref{sec:loogle}.

\subsection{MuSiQue}
\label{sec:musq}

Table~\ref{tab:musicque} presents the results from the MuSiQue multi-hop question-answer dataset. From the table, it can be observed that in the multi-hop QA task, requiring global information: 1) LongLLMLingua can reduce noise in the prompt by eliminating irrelevant information and putting more related information at the beginning or end of the prompt, thereby improving performance by 5.4 points. 2) The performance drop is more pronounced for retrieval-based methods, particularly for n-gram-based methods like BM25. Due to long dependencies, direct matching information is lost, resulting in less relevant information being recalled. 3) The performance of compression-based methods is slightly different. Selective-Context does not distinguish between different modules' sensitivity, resulting in a loss of question and instruction-related information, thereby leading to poorer performance. However, LLMLingua can still retain relevant key information at around a 2x compression ratio. 4) The ablation experiments show that every module designed in LongLLMLingua plays a role in the multi-hop task. The removal of the question-aware coarse-grained and w/ $p(\mathbf{x}_{k}^{\text{doc}}|x^{\text{que}, \text{restrict}}_i)$ modules, which have difficulty in perceiving the importance distribution of corresponding questions, can cause a drop of up to 8 points. Removing the restrict prompt in the question-aware coarse module can also cause a 2-point drop due to the hallucination issue of small LLM. In addition, removing question-aware fine-grained, dynamic compression ratio, and document reordering can all cause a drop of 0.5-2.8 points. 5) Moreover, if the small language model in LongLLMLingua is replaced with GPT2-small, it can further improve the acceleration ratio and still achieve a result that is 2.6 points better than the original prompt.

\subsection{LooGLE}
\label{sec:loogle}

Table~\ref{tab:result_loogle} presents the experiment results in the LooGLE long dependency benchmark, which features longer prompts ($\sim$30k) and more global dependencies. From the table, we can observe that: 1) LongLLMLingua can effectively improve the performance of long context tasks by compressing prompts, even for long dependency tasks. The results show that LongLLMLingua significantly improves performance in tasks such as retrieval, timeline reorder, and computation, with the maximum improvement reaching 15.9 points. 2) The document reorder in LongLLMLingua is effective in all types of tasks, even in tasks highly related to the timeline, it can effectively improve performance by alleviating the "lost in the middle" issue. 3) Retrieval-based methods tend to lose performance in tasks that have longer dependencies, such as computation and reasoning. 4) For compression-based methods, due to the difficulty in perceiving question information, there tends to be a larger performance loss in retrieval tasks within long contexts.

\section{Economic Cost}
\label{sec:econmic_cost}

Table~\ref{tab:cost} presents the estimated per 1,000 samples inference costs for various datasets, encompassing input prompts and generated output text, based on GPT-3.5-Turbo pricing\footnote[19]{https://openai.com/pricing}. Our approach demonstrates substantial savings in computational resources and monetary expenses, particularly in long context situations. Cost reductions of \$3.3 (71.7\%), \$28.5 (90.5\%), \$27.4 (89.5\%), \$2.0 (52.6\%), and \$88.0 (94.0\%) per 1,000 samples are observed for Multi-document QA, LongBench, ZeroScrolls, MuSiQue, and LooGLE, respectively.

\begin{figure*}[htb]
    \begin{tcolorbox}
    \textit{\textbf{Ours w/o Token-level Question-aware:}} \\
    \textbf{Compressed Prompt:} \\
    \textcolor{blue}{\textit{Write a high-quality answer for the given question using only the provided search results (some of which might be irrelevant).}}\\\textcolor{blue}{Document [1](: Physics)}gen,, who received2K, which is ,73,0 in0.  Johnen only to twice6. Mariaie won, for.g was, until1estate he. Two:Mayer (1963). As of 2017, the prize has been awarded\\\textcolor{blue}{\textit{Question: who got the first nobel prize in physics}}\\\textcolor{blue}{\textit{Answer:}}\\
    \textbf{LLMs' Response:} \\
    No answer found in the given search results. 
    \tcblower
    \textit{\textbf{Ours w/ Token-level Question-aware:}} \\
    \textbf{Compressed Prompt:} \\
    \textcolor{blue}{\textit{Write a high-quality answer for the given question using only the provided search results (some of which might be irrelevant).}}\\\textcolor{blue}{1Title: List of Nobelates in} The first Nobel Prize was1 to \boxed{\textcolor{blue}{\textbf{Wilhelmrad}}}, of who received 1582 which,70 in0 en the prize. Skska also won two Nobeles for physics3g01, theate he women prize:ertMayer (1963). As of 2017, the prize has been awarded\\\textcolor{blue}{\textit{Question: who got the first nobel prize in physics}}\\\textcolor{blue}{\textit{Answer:}}\\
    \textbf{LLMs' Response:} \\
    Wilhelmrad\\
    \textbf{LLMs' Response after Subsquence Recovery:} \\
    Wilhelm Conrad Röntgen\\
    \textbf{Ground Truth:} \\
    Wilhelm Conrad Röntgen
    \end{tcolorbox}
    \caption{Comparing the compressed prompt and LLMs' response before and after using Question-aware Fine-grained Compression and Subsequence Recovery($1/\tau$ = 30x, high compression ratio setting) from NaturalQuestions Multi-document QA~\citep{liu2023lost} using GPT-3.5-Turbo.}
    \label{fig:case_token_level_question_aware}
\end{figure*}

\section{Ablation Analysis}
\label{sec:ablation_case}

Figure~\ref{fig:case_token_level_question_aware} illustrates the compressed prompts from the Multi-document QA dataset, comparing the use of contrastive perplexity at a high compression ratio (30x). It shows that without question-aware token-level prompt compression, LongLLMLingua tends to compress key information, a tendency that becomes more pronounced at higher compression ratios. Conversely, employing contrastive perplexity allows for better detection of key information related to the question within the context, thus preserving key information within the compressed prompt.

\section{Cases Study}

Figures~\ref{fig:case_multi_doc_qa}, ~\ref{fig:case_longbench_code}, and ~\ref{fig:case_longbench_trec} display the outcomes before and after compression, as well as the LLMs' responses in various scenarios.

\begin{figure*}[htb]
    \begin{tcolorbox}
    \textbf{Original Prompt:} \\
    ...\\
    \textcolor{blue}{Document [1](Title: Dancing on Ice) }It was confirmed on 25 January 2018, that Dancing on Ice had been recommissioned for an eleventh series to air in \boxed{\textcolor{blue}{\textbf{2019}}}.\\
    ...\\
    \textbf{Compressed Prompt:} \\
    \textcolor{blue}{\textit{Write a high-quality answer for the given question using only the provided search results (some of which might be irrelevant).}}\\\textcolor{blue}{1Title: Dancing on} was confirmed on 2 January 2018 that Dancing on had been recommissioned for an eleventh series air in \boxed{\textcolor{blue}{\textbf{209}}}.\\\textcolor{blue}{Document [2Title: Dan on)} Dan on Ice Dancing on British presented by Phillip Schof alongside Holly Willough from 26 to 2011, and Christine Bleakley from 2012 to 204 The show consists of celebrit and professional partners figure skating in front of a panel of judges The, broadcast on ITV, started on January 2006 and ended on 9 March 2014 after show\' contract not renewed by ITV On 4 September 2017, it was announced that rev series would on I 7 January 201 Sch and Willby returning as a\\ \textcolor{blue}{5(: on} ( on () The third series of a from January to168TV. The from Saturdays, with Holby present Kar,y Sliner Robin Cins returned to Panel", with Ruth H joining the panel as replacement for Natalia Bestova. The commission of the was confirmed by at the07 announcedova depart the series Robinen Bar,ater and Jasoniner announced\\ \textcolor{blue}{7( on} ( )) Dan 2 second of Dan on a from January to1207 ITV It presented Phillip Sch Holly Willough, and judged the "I P consisting Nicky Slater, Nataliaian Karenres Jason Gardiner Karen Barber and Robin Cousins Jaynevill and Christopher Dean co and trained the contestants In this series, cele to ten in first series. The series was won former Kyran Bracken, with Mel Lambert the winner. It announced thatenresge\\\textcolor{blue}{Document []( on Ice} on 08 on TV edition started 8 TV2 The Russian version "\foreignlanguage{russian}{анду}) being on channel0, and renamed in8 to " Ice" (). Its counterpart called "Ice Age (, "Stars on Ice on Channel Oneak IceHviezdyľJ. The Turkish version" is called Dans" ("ance on\\Document1 on Ice its, all,é () and Sje Chris de In series.2 edition\\\textcolor{blue}{](: on Ice world)} Dan Ice is a made competition world format, and been subsequently Italy Chile where titled after series There have a, the show was broadcast on Channel 13 as a\\\textcolor{blue}{Document [17](Title: Dancing on Ice)} the insight to the training of the celebrities over the last week. It was presented by television presenter Ben Shephard and former contestant and "Loose Women" star Coleen Nolan. The show was broadcast from 8 pm to 8.30 pm on Friday evenings on ITV throughout the duration of the main shows season. STV who broadcast the main show did not broadcast this on the Friday evening but after repeating the previous week\'s main show on the following Saturday afternoon. Due to poor ratings, "Dancing on Ice Friday" was axed prior to the 2011 series. The show was based in the\\\textcolor{blue}{\textit{Question: when is dancing on ice on the tv}}\\\textcolor{blue}{\textit{Answer:}}\\
    \textbf{LLMs' Response:} \\
    209 \\
    \textbf{LLMs' Response after Subsquence Recovery:} \\
    2019\\
    \textbf{Ground Truth:} \\
    2019
    \end{tcolorbox}
    \caption{Cases study on NaturalQuestions Multi-document QA dataset~\citep{liu2023lost} in 4x constraint using GPT-3.5-Turbo.}
    \label{fig:case_multi_doc_qa}
\end{figure*}

\begin{figure*}[htb]
    \begin{tcolorbox}
    \textbf{Compressed Prompt:} \\
    \textcolor{blue}{\textit{Please complete the code given below.}}
    \begin{verbatim}
public class MessageArchiveManagement
    private static final long MILLISECONDS_IN_DAY = 24 * 00 *0;
    public static final long_CUP = MCON_DAY
    /.../
          .("",.getStart
          add
 ifget() >0
           Node end("
            end.("
            endNode.Value("", Util.getTimestamp(query.getEnd
addNode
        }        if (.withid null && contact null && !isference
           Node with("           .with           
           .Value("valuewith
           .(
        //    queryMessageive(connection, nextQuery
            final(connectionProtocol(), query
            synchronized (eries)
            //    queries.add(nextQuery } }
    public boolean queryInProgress( contact, OnLoaded 
    moreMessagesLoadedListener)
       ized (eries)
            (Query query : queries)
                if(query.getWith().equals(contact.getUserId()))
    if (query.onMoreMessagesLoaded == null &&MessagesListener 
    null) query.setOnMoreMessagesLoaded(Listener}
                    return true;}} return false;}}
    private void finalizeQuery(Protocol protocol, Query query) {
        synchronized (queries) {
            .remove(query); }
        Contact contact = null;
        if (query.getWith() != null) {
            contact = protocol.getItemByUID(query.getWith()); }
        if (contact != null) {
    \end{verbatim}
    \textcolor{blue}{\textit{Next line of code:}}\\
    \textbf{LLMs' Response:}
    \begin{verbatim}
        contact.setLastMessageTransmitted(query.getEnd());\n
    \end{verbatim}
    \textbf{Ground Truth:} 
    \begin{verbatim}
        if (contact.setLastMessageTransmitted(query.getEnd())) {
    \end{verbatim}
    \textbf{Zero-shot LLMs' Response:} 
    \begin{verbatim}
        contact.removeQuery(query);\n
    \end{verbatim}
    \end{tcolorbox}
    \caption{Cases study on lcc code completion task in LongBench benchmark~\citep{bai2023longbench} in 2,000 constraint using GPT-3.5-Turbo.}
    \label{fig:case_longbench_code}
\end{figure*}

\begin{figure*}[htb]
    \begin{tcolorbox}
    \textbf{Compressed Prompt:} \\
\textcolor{blue}{\textit{Please determine the Type of the question below. Here are some examples of questions.}}\\
\textcolor{blue}{Question:} How is energy created ? \textcolor{blue}{Type} Manner of an action\\
\textcolor{blue}{Question:} What is chocolate ? \textcolor{blue}{Type:} Definition of something\\
\textcolor{blue}{Question:} What is a bone marrow transplant ? \textcolor{blue}{Type:} Definition of something\\
\textcolor{blue}{Question:} What is fear of odors , body , ? \textcolor{blue}{Type} Disease and medicine\\
\textcolor{blue}{Question:} What was the Vietnam War ? \textcolor{blue}{Type:} Definition of something\\
\textcolor{blue}{Question:} was education system in 16s ? \textcolor{blue}{Type:} Other entity\\
\textcolor{blue}{Question:} What is IP address ? \textcolor{blue}{Type:} Definition of something\\
\textcolor{blue}{Question:} are the differences in Catholic Methodist religions ? \textcolor{blue}{Type} of something\\
...\\
\textcolor{blue}{Question:} When was San fire ? \textcolor{blue}{:}  Date\\
\textcolor{blue}{Question:} CNN began broadcasting in what year ? \textcolor{blue}{Type:} Date\\
\textcolor{blue}{Type:} Manner of an action\\
\textcolor{blue}{Question:} What the l behind the ir in the eye called ? \textcolor{blue}{Type} Equ term\\
\textcolor{blue}{Type:} Date\\\
\textcolor{blue}{Question:} What the former name of Zimbabwe ? \textcolor{blue}{Type:} term\textcolor{blue}{Type} something\\
\textcolor{blue}{Question:} What is troilism ? \textcolor{blue}{Type:} Definition of something\\
\textcolor{blue}{:}  What is origin of the word , \textcolor{blue}{Type:} of something\\
\textcolor{blue}{:}  do you name to social security number ? \textcolor{blue}{Type} Manner of an action\\
\textcolor{blue}{:}  that of an employee Universal and Export ? \textcolor{blue}{Type} Individual\\
\textcolor{blue}{:}  anesthetic did Queen Victoria allow to be for the birth of her seventh , in 183 ? \textcolor{blue}{Type:} Disease and medicine\\
\textcolor{blue}{:}  Where isyer 's rock ? \textcolor{blue}{Type} location\\
\textcolor{blue}{Question:} What isymnophobia ? \textcolor{blue}{Type:} Definition of something\\
...\\
 \textcolor{blue}{Type} burns the most calories ?\\
\textcolor{blue}{Type} Sport\\
\textcolor{blue}{:}  In what book I find story of Aladdin ? \textcolor{blue}{Type} In, book and piece an have sex ?\\
\textcolor{blue}{Type:} Manner of an action\textcolor{blue}{:}  What is the acron for rating forer ?\\
\textcolor{blue}{Type} Abbreviation\\
\textcolor{blue}{:}  are the Baltic States ? \textcolor{blue}{Type:} Definition of something\\
\textcolor{blue}{:}  What is appearance , that violates the standards of sexual mor ? Type\\
\textcolor{blue}{:}  Where did the May people live ? \textcolor{blue}{:}  location\\
\textcolor{blue}{:}  What population Kansas ? \textcolor{blue}{Type} number\\
\textcolor{blue}{:}  was the hurr ? \textcolor{blue}{Type:} Event\\
\textcolor{blue}{:}  's a score aymnast exercise ? \textcolor{blue}{Type:} number\\
\textcolor{blue}{:}  year become a state ? \textcolor{blue}{Type:} Date\\
 do go school ? \textcolor{blue}{Type} Reason\\
...\\\textcolor{blue}{\textit{Question: What is a fuel cell ?}}\\\textcolor{blue}{\textit{Type:}}\\
    \textbf{LLMs' Response:} \\
    Definition of something\\
    \textbf{LLMs' Response after Subsquence Recovery:} \\
    Definition of something\\
    \textbf{Ground Truth:} \\
    Definition of something
    \end{tcolorbox}
    \caption{Cases study on trec few-show learning in LongBench benchmark~\citep{bai2023longbench} in 2,000 constraint using GPT-3.5-Turbo.}
    \label{fig:case_longbench_trec}
\end{figure*}

\end{document}